\pdfoutput=1
\documentclass[11pt]{article}

\usepackage[final]{style/acl}
\usepackage{times}
\usepackage{latexsym}
\usepackage{adjustbox}  % in preamble
\usepackage{subcaption}
\usepackage[T1]{fontenc}
\usepackage[utf8]{inputenc}

\usepackage[normalem]{ulem}
\usepackage{ amssymb}
\usepackage{booktabs}
\usepackage{multirow}
\usepackage{enumitem}
\usepackage{colortbl}
\usepackage{bbding}
\usepackage{graphicx}
\usepackage{threeparttable}
\usepackage{hyperref}
\hypersetup{colorlinks=true, urlcolor=magenta}
\usepackage{array}
\usepackage{inconsolata}
\usepackage{microtype}
\usepackage{soul}
\usepackage{url}
\usepackage{amsmath}
\usepackage{amsthm}
\usepackage{booktabs}
\usepackage{algorithm}
\usepackage{algorithmic}
\usepackage[switch]{lineno}
\usepackage{comment}
\usepackage{soul}

\title{From Grounding to Manipulation: Case Studies of Foundation Model Integration in Embodied Robotic Systems}

\author{
Xiuchao Sui$^1$\thanks{Equal contribution. Corresponding authors. \texttt{\{Sui\_Xiuchao, Tian\_Daiying\}@a-star.edu.sg}} \and
Daiying Tian$^1$\footnotemark[1] 
\and Qi Sun $^2$
\and Ruirui Chen$^1$ 
\and \\
\bf{Dongkyu Choi} $^1$ \and
\bf{Kenneth Kwok} $^1$ \and
\bf{Soujanya Poria} $^2$\\
$^1$IHPC, Agency for Science, Technology and Research, Singapore\\
$^2$Nanyang Technological University, Singapore
}
\begin{document}
\maketitle

\begin{abstract}
Foundation models (FMs) are increasingly applied to bridge language and action in embodied agents, yet the operational characteristics of different integration strategies remain under-explored---especially for complex instruction following and versatile action generation in changing environments. We investigate three paradigms for robotic systems: end-to-end vision-language-action models (VLAs) that implicitly unify perception and planning, and modular pipelines using either vision-language models (VLMs) or multimodal large language models (MLLMs). Two case studies frame the comparison: instruction grounding, which probs fine-grained language understanding and cross-modal disambiguation; and object manipulation, which targets skill transfer via VLA finetuning. Our experiments reveal trade-offs in system scale, generalization and data efficiency. These findings indicate design lessons for language-driven physical agents and point to challenges and opportunities for FM-powered robotics in real-world conditions.
\end{abstract}

% Introduction ==========================================================
\section{Introduction}
\label{sec:intro}

% P1 
Natural language is emerging as a universal interface for embodied robotics. Advances in foundation models (FMs) enable robots to follow free-form instructions across perception, reasoning, and motor control, offering the promise of \emph{language-grounded autonomy}. These models include vision–language models (VLMs) \cite{groundingdino,ravi2024sam,ren2024grounded,blip}, multimodal large language models (MLLMs) \cite{llamaVL,Qwen,DS}, and vision–language–action (VLA) models \cite{kim2024openvla,zheng2024tracevla,qu2025spatialvla,bulearning}.

\begin{figure}[t]
    \centering
    \includegraphics[width=\columnwidth]{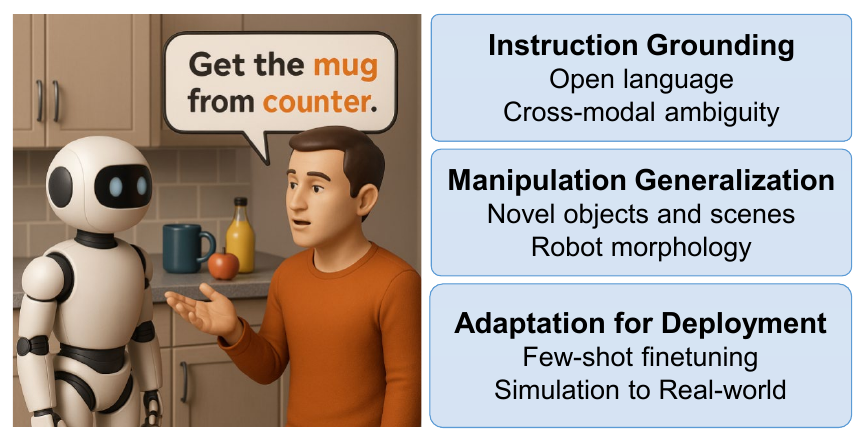}
    %\caption{Challenges in multi-functional robotic system.}
    \caption{Key challenges of FMs in embodied robotics.}
    \label{fig1}
    \vspace{-10pt}
\end{figure}

% P2
However, realizing the promise of \emph{language-grounded autonomy} in deployable systems remains highly challenging. Robots must (i) map ambiguous instructions to the physical world (\emph{instruction grounding}), (ii) execute reliably across novel objects, scenes, and morphologies (\emph{generalizable execution}), and (iii) achieve these goals with limited data (\emph{efficient adaptation}). How effectively different FM integration strategies address these competing requirements remains under-explored (Fig.~\ref{fig1}).

% F1 

%P3
This work offers an empirical study of three integration paradigms: end-to-end VLAs mapping language and vision to actions, MLLM agents orchestrating perception and control, and modular VLM pipelines pairing perception-specialist FMs with task-specific planners (Fig.\ref{fig2}; Table\ref{tab1}).

We evaluate these paradigms through tabletop case studies that expose their complementary strengths and limitations. Two task categories are considered: complex instruction grounding, probing fine-grained understanding and cross-modal disambiguation (Sec.\ref{sec:3}); and object manipulation, assessing skill transfer after VLA fine-tuning under distribution shifts, complemented by comparative and ablation analyses (Sec.\ref{sec:4}).

Our instruction grounding experiments reveal distinct trade-offs across integration strategies. VLM pipelines emphasize interpretability and efficiency, but fall short of peak performance.  They struggle with complex instruction grounding yet achieve moderate object grounding—using less than 1\% of the parameters required by MLLMs. In contrast, MLLMs generalize better on complex instructions but incur substantially higher inference costs, with smaller reasoning-focused models at times outperforming larger ones. VLAs, with tightly coupled perception-to-action pathways, streamline action generation but make it difficult to isolate and assess grounding performance. To probe this gap, we evaluate their perception heads as reference points, which also surfaces open questions for VLA architecture design.

While MLLMs perform well on grounding tasks, their scale limits practical deployment on robotic platforms. Quantization is often adopted as a simple remedy, yet instruction-dependent behaviors emerge, with certain reasoning capabilities degrading disproportionately under naïve compression. This underscores the need for fine-grained quantization strategies. Taken together, our findings clarify the trade-offs between model scale and performance and provide practical guidance for building robotic systems under real-world constraints.

Our object manipulation experiments investigate the skill-adaptation capabilities of VLAs in both real-world and simulated settings. We examine how these models transfer manipulation skills under distribution shifts, evaluating their training dynamics, generalization, and robustness. Results reveal fragile training and slow adaptation in generalist policies, such as OpenVLA \cite{kim2024openvla}, compared to task-specific counterparts, underscoring the challenges of real-world deployment.”

 % Our  To assess generalization, we analyze robustness to environmental perturbations and to variation in scene complexity from object arrangements, which remain particularly demanding for robots operating in real-world.

\begin{comment}
% P5
%While fine-tuning for manipulation (Sec.~\ref{sec:4}) evaluates generalization performance, we hope these case studies offer practical insights for developing robotic systems with foundation models (FMs). We categorize leading VLA models into autoregressive and diffusion-based approaches based on their action generation mechanisms, and compare their performance through fine-tuning. Futhermore, we evaluate the impact of action chunking, examine robustness to distraction and assess performance across both simulated and real-world tasks.    
\end{comment}

% P6
In summary, our main contributions are:
\begin{itemize}[leftmargin=1em, itemsep=0pt, topsep=0pt]
\item We analyse three FM integration paradigms on shared embodied tasks designed to probe the capabilities and trade-offs of FMs.

\item We release a dataset and code\footnote{\url{https://github.com/xiuchao/InstructionGrounding}} for evaluating instruction grounding and object manipulation, covering cross-modal reasoning and skill adaptation under varied layouts.

\item We providing timely insights into state-of-the-art VLAs and MLLMs,  investigating their capabilities and failure modes and distilling practical trade-offs to guide practitioners in selecting FM stacks for language-driven embodied agents.  

\item We also release a complete end-to-end claw-machine robot system as a real-world FM integration demo\footnote{\url{https://github.com/HRItdy/claw_machine}}.

\end{itemize}

% Table 1 -------------------------
\begin{table*}[htbp]
\begin{minipage}{\textwidth}
\adjustbox{max width=\linewidth}{
\begin{tabular}{@{}lccccll@{}}
\toprule
    \multirow{2}{*}{\begin{tabular}[c]{@{}l@{}} \\ \textbf{Pipelines for} \\ \textbf{Robot Systems}\end{tabular}}  & \multicolumn{3}{c} {\textbf{Instruction Grounding}}   & \multicolumn{2}{l} {\textbf{Manipulation Generalization}}  & \textbf{Adaptation for Deployment}    \\ 
    %\cmidrule(l){2-7} 
    \cmidrule(lr){2-4} \cmidrule(lr){5-6} \cmidrule(lr){7-7}
    & \begin{tabular}[c]{@{}l@{}}Visual \\inputs \end{tabular} 
    & \begin{tabular}[c]{@{}l@{}}Multi-round  \\\ dialogue\end{tabular} 
    & \begin{tabular}[c]{@{}l@{}}CoT \\reasoning \end{tabular} 
    & \begin{tabular}[c]{@{}l@{}}Morphology\\ independent\end{tabular} & Skill sets  & \begin{tabular}[c]{@{}l@{}} Data Efficiency \end{tabular}   \\ \midrule

    End-to-End VLA models       & \Checkmark       & \XSolidBrush    & \XSolidBrush  & \XSolidBrush  & \begin{tabular}[c]{@{}l@{}} Wide range \end{tabular} & Data-hungry finetuning \\
    Modular VLM pipelines  & \Checkmark       & \XSolidBrush     & \XSolidBrush     & \Checkmark    & \multirow{2}{*}{\begin{tabular}[c]{@{}l@{}}Controller- \\ specific\end{tabular}} &  Cheap finetuning  \\
    Multimodal LLMs agents     & \Checkmark       & \Checkmark     & \Checkmark   & \Checkmark   &    & In-context learning\\ \bottomrule
\end{tabular}}
\caption{Comparison of foundation model integration strategies in embodied robotic systems, highlighting differences in instruction grounding, manipulation generalization, and adaptation methods.}
\label{tab1}
\end{minipage}
\end{table*}

% Integration Strategies =============================================
\section{Foundation Model Integration for Language-Guided Robotics}
\begin{figure}[t]
  \centering
  \includegraphics[width=\linewidth]{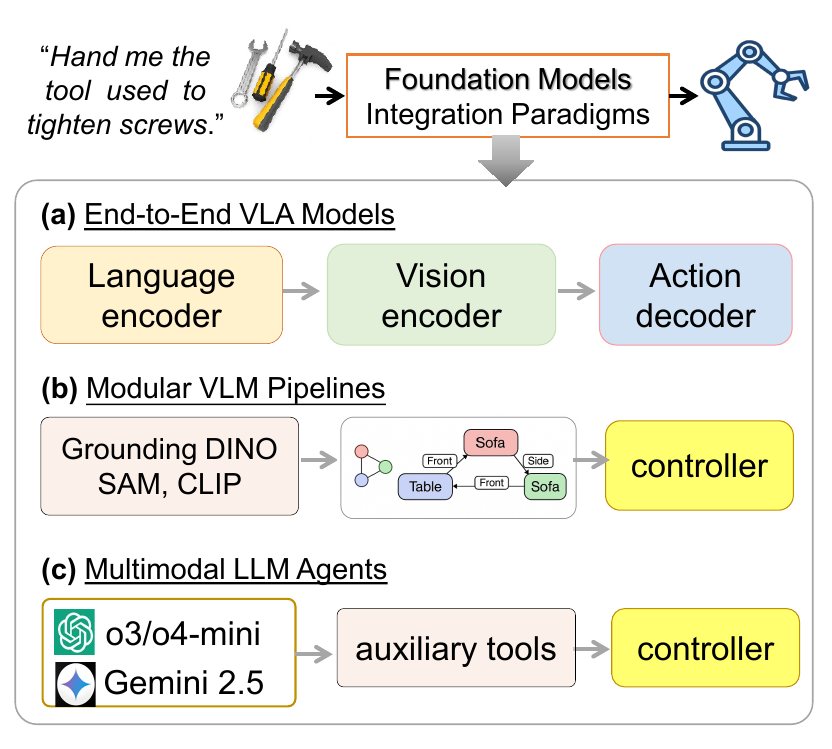}
  \caption{Three FM integration strategies for embodied robotics, highlighting distinct interfaces between language, perception, and control.}
  \label{fig2}
\end{figure}

Concerning how FMs are integrated into robot systems, we identified the following three types of integration strategies (Fig.~\ref{fig2}). In the following, We briefly describe each strategy along with its respective advantages and limitations.

% Figure 2

%\caption{\XC{A taxonomy of FM integration strategies in robotics (a) End-to-end VLA, where a single model directly maps multimodal inputs to action. (b) LLM-based code-generation policy, where the FM outputs executable low-level code. (c) Modular pipelines with specialist VLMs used for perception and grounding. (d) Multimodal LLM agents that often using auxiliary reasoning tools. Each strategy reflects a distinct interface between language, perception, and control.

\begin{comment}
[xc] the previous caption is to long, del for more space.

(a) Vision-Language-Action models directly map instructions and scene inputs to actions. (b) LLM-based Code Policies generate high-level action scripts from instructions. (c) Specialist VLMs provide grounded scene understanding. (d) Multimodal LLMs leverage auxiliary tools for comprehensive perception and reasoning.
\end{comment}

% Sections
\subsection{End-to-End VLA Models} 
\label{sec:2.1}
\paragraph{Definition.} VLAs operate in an end-to-end manner, directly translating visual observations and natural language instructions into low-level actions without decoupled perception, language, and control modules (Fig.~\ref{fig2}a). Two mainstream paradigms have emerged within this framework: auto-regressive and diffusion-based action generation. Through large-scale pretraining, these models acquire broad capabilities that support generalization across tasks. However, efficient adaptation to real-world settings remains a significant challenge.

\paragraph{Autoregressive Models.} These models generate actions step by step, with each action conditioned on the current perceptual input and previously generated outputs. These models typically encode visual observations and language into a shared latent space, then employ a transformer-based decoder to autoregressively predict low-level control tokens, such as joint angles, end-effector displacements and gripper states, which are typically obtained by discretizing continuous motor signals into token sequences. 

RT-1 \cite{brohan2022rt} and RT-2 \cite{brohan2023rt} demonstrated the effectiveness of large-scale pretraining for task generalization. Building on these scaling successes, OpenVLA \cite{kim2024openvla} became a landmark open-source effort, combining Llama-2 with DinoV2 and SigLIP, trained on the large-scale Open-X-Embodiment dataset \cite{o2024open}. Subsequent works such as Emma-X \cite{sun2024emma}, NORA \cite{hung2025nora} and TraceVLA \cite{zheng2024tracevla}, further advanced this line through dataset quality, stronger backbones and improved spatiotemporal prompting.

Subsequent models further incorporate \textit{richer modalities} such as tactile sensing \cite{yang2024binding,zhao2024transferable}, \textit{structural priors} including spatial representations distilled from VLMs \cite{gemini} and 3D spatial relationships in SpatialVLA \cite{qu2025spatialvla}, and refinements to the \textit{embodied latent space}, as in UniVLA \cite{bulearning}, which learns task-specific latent representations.

\paragraph{Diffusion-based Models.} They generate actions by progressively denoising trajectories from noise, modeling the distribution of future actions as a whole rather than step by step. Like autoregressive VLAs, these models encode visual observations and language into a latent representation, but instead apply a diffusion process that iteratively refines action sequences into feasible trajectories. Diffusion-based models trade inference speed for greater trajectory coherence. 

Diffusion Policy \cite{chi2023diffusion} pioneered the use of diffusion models for visuomotor policy learning, showing improved global consistency and robustness compared to autoregressive baselines. Octo \cite{team2024octo} applies conditional diffusion decoding for action sequence prediction. Subsequent models further  scales diffusion heads into larger, dedicated policy modules \cite{reuss2024multimodal,wen2024diffusion,li2024cogact}, and integrate trajectory-level guidance to improve long-horizon planning \cite{fan2025diffusion}. Notable models like  $\pi_0$ \cite{black2024pi_0} combine pretrained MLLM (PaliGemma-3B) with flow-matching-based action experts to produce continuous, high-frequency control across robot embodiments. $\pi_{0.5}$ \cite{intelligence2025pi_} builds on this by co-training across diverse environments and data modalities to improve generalization to new settings.

Beyond architectural advances, diffusion-based models increasingly targets diverse embodiments, extending from single-arm manipulation to bimanual systems and humanoid robots. Representative examples include RDT-1B \cite{liu2024rdt}, DexVLA \cite{wen2025dexvla}, and GR00T N1 \cite{bjorck2025gr00t}, which demonstrate the scalability of FM-driven action generation to more complex morphologies.

\paragraph{Strengths and Limitations.}
\begin{comment}
% raw verson
VLAs provide a unified, end-to-end framework for robotic manipulation, (i) seamlessly integrating visual, language, and action modalities. Leveraging large-scale pretraining and fine-tuning, these models exhibit (ii) strong potential for generalizing across diverse manipulation tasks and robotic embodiments. However, their performance is constrained by the (iii) \emph{limited availability} of high-quality, diverse robotic datasets. Pretraining can also introduce biases from training distributions, leading to (iv) degraded performance in out-of-distribution scenarios, such as \emph{novel tasks} or \emph{different robotic embodiments}. Therefore, despite their potential, further work is needed to enhance their robustness and generalization across real-world settings, for example, efficient adaptation using few-shot data.    
\end{comment}
Leveraging large-scale pretraining, VLAs hold the potential to generalize across diverse manipulation tasks and robot morphologies. Yet progress is constrained by the scarcity of high-quality, diverse robotic datasets, which limits both scale and coverage. Pretraining can also propagate biases from training distributions, causing degraded performance on novel tasks, in unseen environments, or with unfamiliar embodiments. Despite their promise, VLAs remain brittle in real-world deployment, as highlighted in our \emph{Object Manipulation} case study (Section~\ref{sec:4}), underscoring the need for better data curation, bias mitigation, and efficient adaptation strategies such as few-shot and continual learning.

% ---------------------------------------------------------
% Strengths: interpretability, cheap finetuning, plug-and-play with classical control. 
% Weaknesses: pipeline break-points, slower cross-modal reasoning.
% For example, \cite{RSS24demo} developes a sample collection robot that leverages GroundingDINO and SAM to guide its behavior.}

\subsection{Modular VLM Pipelines} 
\paragraph{Definition.} In this paradigm (Fig.~\ref{fig2}b), perception is handled by a  \emph{specialist} VLM that outputs symbolic scene information, typically grounded 2D/3D bounding boxes, segmentation masks, or referring expression pointers.  A downstream planner or policy module then consumes this structured representation to generate low-level actions.  The language channel is therefore \emph{disentangled} from motor control, allowing each module to be tuned independently, thus preserving the transparency and plug-and-play advantages of classical planning.

\paragraph{Representative systems.} Language-promptable specialist VLMs endow modular stacks with zero-shot semantics for various robotics pipelines. \cite{RSS24demo} demonstrates an end-to-end \textit{sample collection robot system} that uses GroundingDINO \cite{groundingdino} to localize objects and refines each box with SAM \cite{ravi2024sam} masks before passing them to classical grasp-and-place controllers, illustrating this paradigm’s practicality in real deployments. \cite{3dSceneGraph} aggregates these modules into a floor–room–object hierarchy, showcasing their usage in long-horizon language-conditioned navigation across multi-story buildings.

\paragraph{Strength and Limitations.}
Modular VLM pipelines strike a balance between transparency and adaptability, and delivers practical benefits:
(i)~\emph{interpretability}, detections can be directly inspected; (ii)~\emph{efficiency}, models typically contain 100M$\sim$600M parameters, only about 1\% $\sim$ 6\% the size of representative 10B-scale MLLMs \cite{llamaVL}. However, they are limited by (i)~\emph{interaction rigidness} compared to more flexible MLLMs, and (ii) \emph{pipeline brittleness} where perception errors propagate without mitigation (Fig.~\ref{fig2}b; Table \ref{tab1}). Their effectiveness depends on robust open-vocabulary grounding—precisely the capability highlighted in our \emph{Instruction Grounding} case study (Section~\ref{sec:3}).

% -------------------------------------------------------
\subsection{Multimodal LLM Agents as Orchestrators}
\label{subsec：2.3}

% LLM-centric architectures that call vision tools and low-level controllers via function calls. 
% strengths:  strong language generalization, flexible tool
% Weaknesses: tool-API engineering.
\paragraph{Definition.} In this paradigm (Fig.~\ref{fig2}c), MLLMs take raw user utterances, selectively invoke vision tools (\emph{e.g.}, a detector or depth estimator) via function calls, reason over their outputs in context, and issue high-level action primitives to a low-level controller. An MLLM agent thus places a large tool-calling language model at the center of the control loop, acting as a \emph{cognitive hub} that binds perception and control through natural language.

\begin{comment}
% raw version (not future proof)
Models such as GPT-4V, Qwen-VL \cite{Qwen}, and DeepSeek-VL \cite{DS} were included, but more recent releases were not covered—reflecting the rapid pace at which new models continue to emerge.    
\end{comment}
\paragraph{Representative Systems.} MLLMs are playing increasingly important roles in robotics. Gemini Robotics \cite{gemini} integrates perception, spatial reasoning, and trajectory synthesis into a single Gemini-2.0 backbone \cite{deepmind2024gemini}, serving as an embodied brain. ManiLLM \cite{maniLLM}, in a similar spirit, leverages the common-sense and reasoning capabilities of MLLMs by fine-tuning adapter modules with a chain-of-thought training paradigm, enabling accurate pose prediction and precise manipulation. These works illustrate the emerging trend of MLLMs shifting toward the role of a \emph{cognitive hub} in robot systems. Hub-LLaMA \cite{hub_llama3.2v} further builds a modular agent-orchestration system for household object management, using LLaMA 3.2 Vision \cite{llamaVL} for open-vocabulary perception to ground task plans, though its limitations are not discussed.  

Somewhat related to our work, \cite{hub_inhouse} evaluates the suitability of MLLMs as a ``brain'' for in-home robotics, providing a benchmark that compares models across perception, visual reasoning, and task planning. The benchmark included a few MLLMs available at the time, while newer releases were not covered—illustrating the rapid pace of progress in this area.

\paragraph{Strengths and Limitations.} MLLMs excel in (i)~\emph{visual commonsense reasoning}, leveraging extensive language priors to generalize to novel concepts beyond the reach of most specialist VLMs, and (ii)~\emph{instruction following} with support for fine-grained visual understanding and dynamic planning. Despite their expressive power, however, these models are (iii) ~\emph{resource-intensive}, posing challenges for deployment---particularly on mobile robotic platforms. We further examine the capability limits and trade-offs of MLLMs (Section~\ref{sec:3}).

%\XC{We next examine the capabilities and limitations of these integration strategies in real-world settings. Section~\ref{sec:3} addresses complex instruction following, requiring integrated perception and reasoning with cross-modal disambiguation. Section~\ref{sec:4} evaluates action generation across diverse VLA systems.}

% Case Studies: VLM & Multimodal LLMs ==============================================
\section{Case Studies on Instruction Grounding}
\label{sec:3}
% Figure 3 --------------------
\begin{figure}[!t]
    \centering
    \includegraphics[width=\linewidth]{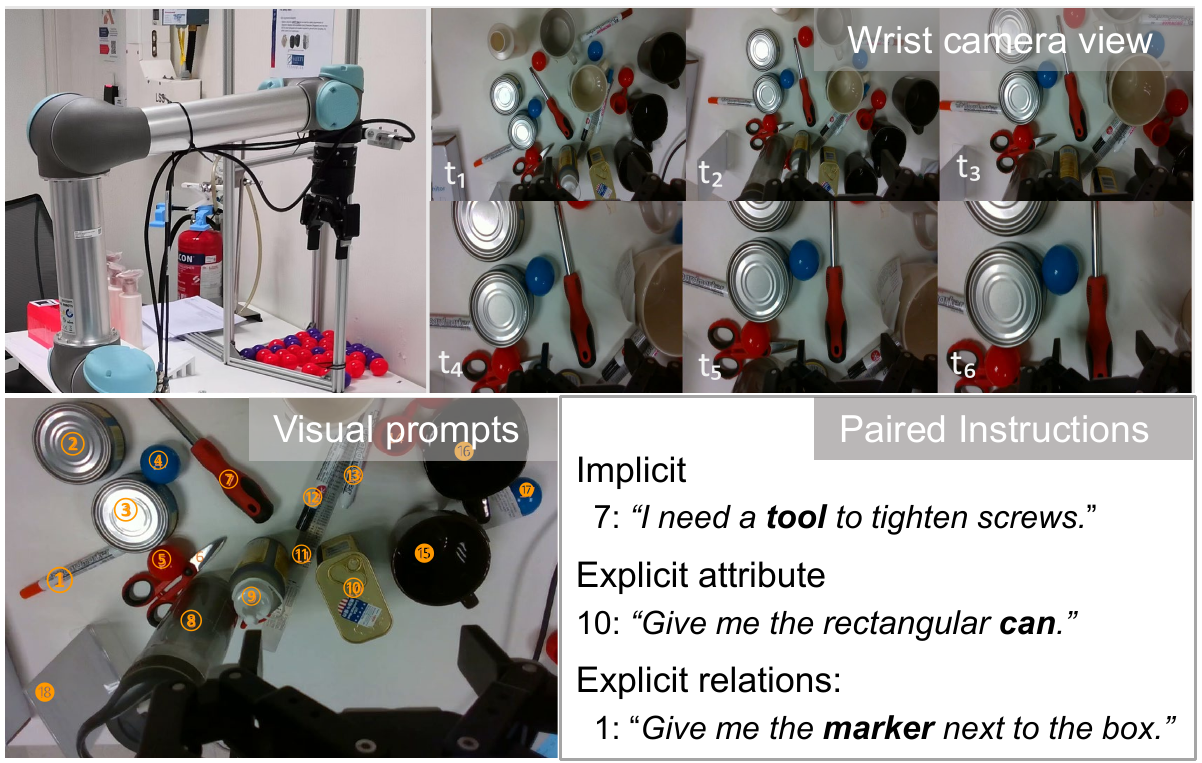}
    %\caption{A case study of cluttered scene manipulation.}
    \caption{Experimental setup for two case studies in a cluttered tabletop environment. The top row shows egocentric video data collected for the manipulation case study. The bottom row is an example setup for the instruction grounding task, including an annotated visual prompt paired with complex instructions in three forms: implicit, explicit with attributes and spatial references.}
    \label{fig3}
\end{figure}

\begin{comment}
Earlier robotics work relied on dense reward functions over low-dimensional state vectors \citep{gupta2018}, or on ResNet-based visual encoders for zero-shot goal cues \citep{Specification1}. FMs broaden this channel: specialist VLMs enable open-vocabulary perception, while multimodal LLMs bring embodied commonsense reasoning, allowing non-experts to instruct robots in plain language. However, open language is inherently ambiguous---different linguistic parses can map to conflicting visual referents (Fig.~\ref{fig3}). Robust instruction grounding thus depends on how effectively an FM can ground user instructions to visual scenes while resolve \emph{cross-modal ambiguities} while remaining computational-efficiency.

Natural language \emph{instruction grounding} translates a user’s intent into an unambiguous, actionable goal within the scene. This section presents an up-to-date evaluation of \emph{instruction grounding}, examining models' abilities to ground open-vocabulary objects and complex instructions. While model quantization reduces computational demands and aids deployment, its performance trade-offs remain underexplored; our case study offers quantitative insights and practical guidance. Additionally, the rapid pace of model development---e.g., LLaMA \cite{meta2025llama4}, Gemini \cite{deepmind2025gemini}, o3, and o4 \cite{openai2025o3o4}---introduces both new capabilities and evaluation challenges, particularly in embodied manipulation settings.
\end{comment}
Natural language \emph{instruction grounding} translates user intent into actionable goals within a visual scene. Our case study evaluates grounding performance under challenging cross-modal disambiguation, highlighting trade-offs and offering guidance for efficient deployment.

\paragraph{Benchmark Dataset.}
To isolate grounding ability from general vision priors, we design carefully controlled scenarios using household objects placed on a tabletop. These objects are widely represented in FM training corpora, while the tabletop setup minimizes variation in lighting and camera angles—ensuring the evaluation primarily reflects grounding performance.

We curated a new Instruction Grounding benchmark (Fig.~\ref{fig3}) consisting of images with multiple household objects, each tagged with numbers as visual prompts and paired with instructions targeting visual commonsense and cross-modal disambiguation. For example, “\emph{pick up the red-capped marker}” requires attribute reasoning to choose among markers, while “\emph{grasp the cup in front of the screwdriver}” tests spatial reasoning (Appendix Table~\ref{tab:A3}).  These tasks highlight how grounding errors can lead to execution failures in embodied systems.

% Table 2 ----------------------
\begin{table}[t]
\centering
\adjustbox{max width=\linewidth}{
\begin{tabular}{lcccc}
\toprule
\textsc{Model}  & \textsc{Easy}   & \textsc{Medium}  & \textsc{Hard}  & \textsc{Avg} \\
\midrule
\multicolumn{5}{c}{\textbf{Specialist VLMs}} \\
GroundingDINO-86M  & \text{0.518}  &\text{0.357}  & \text{0.349}  &\text {0.408} \\
GroundingDINO-145M  & \text{0.443}  &\text{0.320}  & \text{0.355}  &\text {0.372} \\

\\
\multicolumn{5}{c}{\textbf{Proprietary MLLMs}} \\
Gemini2.5-Pro-Exp  & \textbf{0.904}  &\text{0.765}  & \textbf{0.793}  &\text {0.821} \\
Gemini2.0-Flash  & \text{0.884}  &\text{0.738}  & \text{0.678}  &\text {0.767} \\
GPT-5-auto  & \text{0.881}  &\textbf{0.829}  & \text{0.760}  &\textbf {0.823} \\
GPT-5-mini  & \text{0.749}  &\text{0.737}  & \text{0.776}  &\text {0.754} \\
%GPT-4.5 & \text{0.837}  &\text{0.723}  & \text{0.739}  &\text {0.766} \\
GPT-4o  & \text{0.814}  &\text{0.745}  & \text{0.683}  &\text{0.747} \\
GPT-4o-mini  & \text{0.803}  & \text{0.722} & \text{0.604} & \text{0.710} \\
o4-mini  & \text{0.721}  & \text{0.769} & \text{0.710} & \text{0.733} \\
%GPT-4V            & 0.470     & 0.476    & 0.467   & 0.471 \\
\\
\multicolumn{5}{c}{\textbf{Open-source MLLMs}} \\
Llama-3.2V-90B   & 0.722    & \textbf{0.701}   &\textbf{0.657}    & \textbf{0.693} \\
%Llama-3.2V-90B-Q4 & 0.664    & 0.542    & 0.575   & 0.594 \\
Llama-3.2V-11B    & 0.583    & 0.569    & 0.547   & 0.566 \\
%Llama-3.2V-11B-Q4 & \underline{0.745}    & 0.563    & 0.549   & 0.619 \\
Llama-4-Maverick & 0.698    & 0.576    & 0.634   & 0.636 \\
Llama-4-Scout  & \textbf{0.776}    & 0.615    & 0.624   &0.672 \\
Qwen2-VL-72B     & 0.686     & 0.614    & 0.558   & 0.619 \\
Gemma-3-27B  & 0.452    & 0.384    & 0.267   & 0.368 \\
DS-Janus-Pro-7B & 0.444 & 0.330 & 0.317 & 0.364\\
Phi-3.5-Vision-4.2B & 0.291  & 0.357    & 0.205   & 0.284 \\
\\
\multicolumn{5}{c}{\textbf{Base MLLMs in VLAs}} \\
PaliGemma-3B ($\pi_0$) & \text{0.118}  & \text{0.07}    & \text{0.05}  & \text{0.079} \\
QwenVL-3B (NORA)  & \textbf{0.519}  & \textbf{0.535}    & \textbf{0.577}   & \textbf{0.543} \\

\bottomrule
\end{tabular}
}
\caption{Object grounding performance of specialist VLMs and MLLMs across cluttered scenes of varying complexities, with macro accuracy reported.}

\label{tab:2}
\end{table}

% Figure 4 ----------------------
\begin{figure*}[t]
  \includegraphics[width=0.48\linewidth]{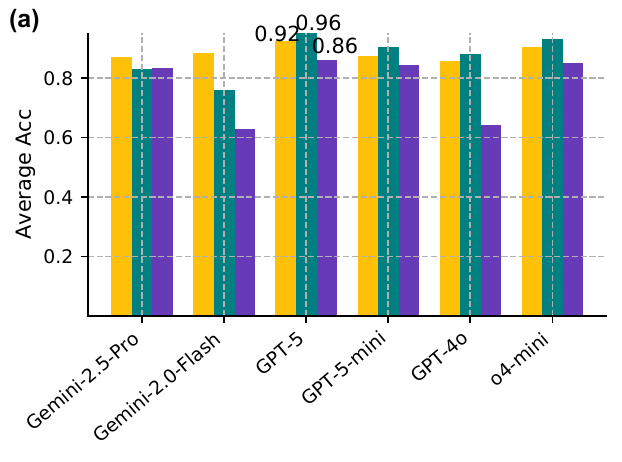} \hfill
  \includegraphics[width=0.48\linewidth]{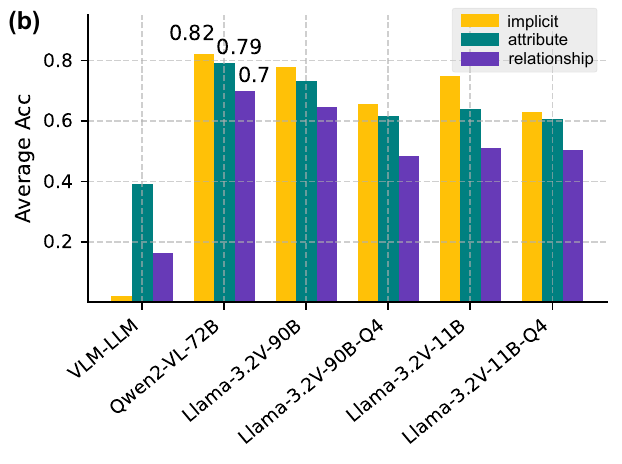}
  \caption {Performance of complex instruction grounding across modular VLM pipeline and MLLMs. Macro accuracy is reported across instruction types---implicit, attribute-based, and relationship-based. Subfigures show (a) proprietary models and (b) open-source models along with their Int4-quantized variants.}
  \label{fig4}
\end{figure*}

% --------------------------------
\paragraph{Zero-Shot Object Grounding.}
\begin{comment}
% raw
We begin with a foundational question for instruction grounding: \textit{Can FMs accurately recognize objects in cluttered open scenes?} Table \ref{tab:2} presents the performance hierarchy for specialist VLMs and a range of MLLMs, serving as a basis for deeper analysis of ambiguity resolution in later sections.
\end{comment}

We start with a necessary first question for instruction grounding: \textit{Can FMs reliably recognize objects in cluttered, open-world scenes?} Table~\ref{tab:2} highligh key observations and point to opportunities for VLA design.

\begin{itemize}[leftmargin=1em, itemsep=0pt, topsep=0pt]
    \item Despite their popularity in modular pipelines, GroundingDINO achieves around 0.3-0.4 accuracy, as it struggles with featureless objects, e.g. \textit{`the can'}. Moreover, it is brittle in open scenes, e.g. a \textit{`screwdriver'} is constantly recognized as a \textit{`marker'}, which instead is an easy case for MLLMs which embodied large volume of visual commonsense (Appendix Fig.~\ref{fig12}). 
    
    \item Gemini~2.5-Pro and GPT-5 achieve top performance with an average score of 0.82, followed by Gemini~2.0-Flash and GPT-4.5. Open-source models continue to lag behind, with LLaMA~3.2-Vision~90B reaching 84\% of the performance of top proprietay models.

    \item Small- and medium-scale models (e.g., Gemma-27B, Phi-Vision) generally fall below the specialist threshold, underscoring their limitations for fine-grained object grounding in open scenes. Notable exceptions include GPT5-mini (0.754) and QwenVL-3B (0.543), which offer favorable speed–accuracy trade-offs, achieving reasonable performance without high API costs or heavy memory footprint of larger models.
    
    \item PaliGemma-3B in $\pi_0$ struggles to follow structured prompts and can only follow simpler instructions. Its grounding capability is highly limited, often detecting just a single object per scene. Reported accuracy is therefore approximate, based on a looser evaluation criterion.  While QwenVL-3B adopted in NORA, reaches higher grounding accuracy, which may partially explain its stronger performance on out-of-distribution manipulation tasks (Appendix Table~6).

    \item Two \textit{design questions} emerges: (i) how should perception modules be selected or adapted to improve instruction-following in embodied systems? and (ii) can compact, grounding-capable perception modules be distilled from large-scale models to support efficient deployment? %The perception base MLLMs contribute to VLAs performance in manipulation was demonstrated \cite{gemini}, while remain unclear in open-source community.

\end{itemize}

% -------------------------------------
\paragraph{Zero-Shot Complex Instruction Grounding.} 
This task is framed as a multiple-choice problem, where the model selects the correct object index in a cluttered scene given three types of natural language instructions: implicit, attribute-based, and relationship-based. Each type probes a distinct grounding challenge (Fig. \ref{fig4}). As a baseline, we evaluate a modular VLM pipeline in which GPT-4 parses the instruction to infer likely targets, queries GroundingDINO to detect candidate objects, and selects from the detected boxes—essentially guessing without directly perceiving the scene.

\begin{itemize}[leftmargin=1em, itemsep=0pt, topsep=0pt]
    \item \textit{Implicit Instruction Grounding}. Instructions like “I need a tool to tighten the screws” only refer to the target object implicitly, and the model needs to infer the target object using its common sense priors. For such instructions, the modular VLM pipeline struggles to select a screwdriver, lacking embedded affordance reasoning. In contrast, MLLMs perform well, reflecting strong visual commonsense. GPT-5 achieves 0.92 accuracy, while GPT-4.5 demonstrates exceptional performance (0.94), though its high inference cost---$20\times$ that of Gemini 2.5 makes it cost-prohibitive for most applications (Appendix Table ~\ref{tab:5}).

    \item \textit{Relational Reasoning Remains Challenging.} This category requires resolving referential ambiguity through implicit chain-of-thought reasoning: grounding objects, modeling spatial relationships, and disambiguating targets (e.g., identifying the correct mug among many based on “next to something”). Accuracy drops significantly nearly across all models. GPT-5, Gemini 2.5-Pro and o4-mini achieve accuracy above 0.80---demonstrating the benefits from embodied training data and strong reasoning capabilities. Notably, o4-mini is a medium-sized model, yet it outperforms larger models like GPT-4o on relational instructions---suggesting that structured reasoning may help close, or even overcome the performance gap brought by different model scales.

    %\item  \XC{\textit{Open-Source Progress.} Qwen2-VL-72B outperforms best across all instruction types (LLaMA-4 series did not surpass its predecessor in object grounding, we focus on LLaMA-3 models). While small open-source models remain limited in handling complex grounding tasks, LLaMA-3.2-11B achieves performance comparable to GPT-4o-mini. }

    \item \textit{Instruction-Dependent Quantization Effects.} INT4 quantization reduces the model size by over 70\%, making it an attractive choice for deployment. In Llama 3.2 Vision, we observe that it disproportionately impacts implicit and relational instruction grounding, indicated by the relative accuracy drop of $14\%-17\%$, while attribute grounding is more robust with only 4\% loss. Despite reduced precision, quantized 11B models offer a speed–accuracy balance for low-resource settings.
    Our findings underscore the need for \emph{fine-grained quantization strategies} that preserve the most important high-level reasoning capabilities under resource constraints. 
    
\end{itemize}

%For VLA, we did not probe their capacity for instruction grounding through tasks involving open-ended, ambiguous commands-Why?.  a very short paragraph.

% Table 3 ---------
\begin{table*}[ht]
\centering
\small % Reduce font size
\setlength{\tabcolsep}{4pt} % Reduce column padding
\renewcommand{\arraystretch}{1.1} % Adjust row spacing
\begin{tabular}{lccccc}
\toprule
\textsc{Models} & \textsc{LIBERO-Spatial} & \textsc{LIBERO-Object} & \textsc{LIBERO-Goal} & \textsc{LIBERO-Long} & \textsc{Average} \\
\midrule
OpenVLA finetuned & 84.7 & 88.4 & 79.2 & 53.7 & 76.5 \\
$\boldsymbol{\pi_0}\ $\textbf{finetuned} & \textbf{96.8} & \textbf{98.8} & \textbf{95.8} &	\textbf{85.2} & \textbf{94.15} \\
$\pi_0\text{-FAST}$ finetuned &96.4 &	96.8 & 88.6 & 60.2 & 
 85.5\\ 
SpatialVLA finetuned-AC & 88.2 & 89.9 & 78.6 & 55.5 & 78.1 \\
NORA-finetuned & 85.6 & 87.8 & 77.0 & 45.0 & 73.9 \\
NORA-finetuned-AC & 85.6 & 89.4 & 80.0 & 63.0 & 79.5 \\
NORA-Long-finetuned & 92.2 & 95.4 & 89.4 & 74.6 & 87.9 \\
\bottomrule
\end{tabular}
\caption{Success rates (\%) on the LIBERO Simulation Benchmark across four task suites, each evaluated over 500 trials. Results for SpatialVLA are from \cite{qu2025spatialvla}; Results for $\pi_0$ are from \cite{black2024pi_0}, using pretrained models on LIBERO benchmarks. “AC” denotes the use of action chunking. The comparison in the Appendix highlights its impact on performance. The finetuned $\pi_0$ model achieves the highest performance.}
\label{tab:4}
\end{table*}

% VLA Case Studies ===========================
\section{Case Studies on Robotic Manipulation}
\label{sec:4}
%  We evaluate their adaptability through fine-tuning under distribution shifts, reflecting real-world deployment demands. To assess generalization, we analyze robustness to environmental perturbations and variation in object appearance and robot morphology. 

%In this section, we evaluate the performance of autoregressive and diffusion-based VLA models through a set of experiments. In particular, we conduct both partial fine-tuning and full fine-tuned, reporting success rates across various tasks. We examine model robustness and analyze the performance gap between simulation and real-world settings. The results are presented and discussed. More detailed results can be found in the Appendix.

Now we shift the focus to \emph{skill adaptation}. In an ideal deployment scenario, a pretrained VLA---already endowed with broad visuomotor skills---should be retargeted to a new manipulation task with minimal data and fast convergence. We use fine-tuning, the standard practice for adaptation, as a probing lever to evaluate how the state-of-the-art VLA models adapt to new tasks and deployment conditions.

Given the scale of VLAs, we compare \textbf{partial fine-tuning}, which leverages our curated probing dataset (Fig.~\ref{fig3}) and its inherent distribution bias to study convergence behavior, and \textbf{full fine-tuning}, which uses large-scale datasets to minimize the training loss. Our evaluation focuses on three key aspects: (i)~\emph{training dynamics}---how quickly and smoothly training converges; (ii)~\emph{generalization}---how well the resulting policies perform on various tasks; and (iii)~\emph{robustness}---how well the resulting policies handle environmental distractors. Our experiments highlight the performance of VLA models in different settings, offering practical suggestions for practitioners who have to adapt large VLAs under tight data, time and compute budgets.

\paragraph{Real-World Skill Adaptation.} 
\begin{comment}
% tdy raw
Our fine-tuning process consists of two stages: (1) To assess convergence behavior under distribution shift, we collected a custom dataset (Appendix~\ref{app:data}) with a distribution bias relative to common pretraining datasets included in the Open-X-embodiment \cite{o2024open} and LIBERO datasets \cite{liu2023libero}. We used it to partially fine-tune several recent VLA models and trained Diffusion Policy (DP) and Action Chunking Transformer (ACT) from scratch. The results are shown in Fig.~\ref{fig5}; (2) For full fine-tuning, we leveraged larger benchmark datasets, Open-X-embodiment and LIBERO, to fully fine-tune RT-1, OpenVLA, SpatialVLA and NORA, and compared their performance. The results are shown in Fig.~\ref{fig:6}.
\end{comment}

Our fine-tuning process consists of two stages: (1) To assess convergence on tasks beyond the coverage of common pretraining datasets, we collected a custom dataset (Appendix~\ref{app:data}) focused on screwdriver-picking in cluttered tabletop scenes. This task introduces both object- and scene-level distribution gaps, as screwdrivers and dense clutter are largely absent from Open-X-Embodiment \cite{o2024open}. We used this dataset to partially fine-tune generalist VLA models, and to train compact task-specific models such as Diffusion Policy (DP) and Action Chunking Transformer (ACT) from scratch. (2) For full fine-tuning, we leveraged larger benchmarks—Open-X-Embodiment and the simulated LIBERO dataset \cite{liu2023libero}—to fully fine-tune RT-1, OpenVLA, SpatialVLA, and NORA, and compared their performance.

% Figure 5 ----------
\begin{figure}[t]
    \centering
    \includegraphics[width=0.96\columnwidth]{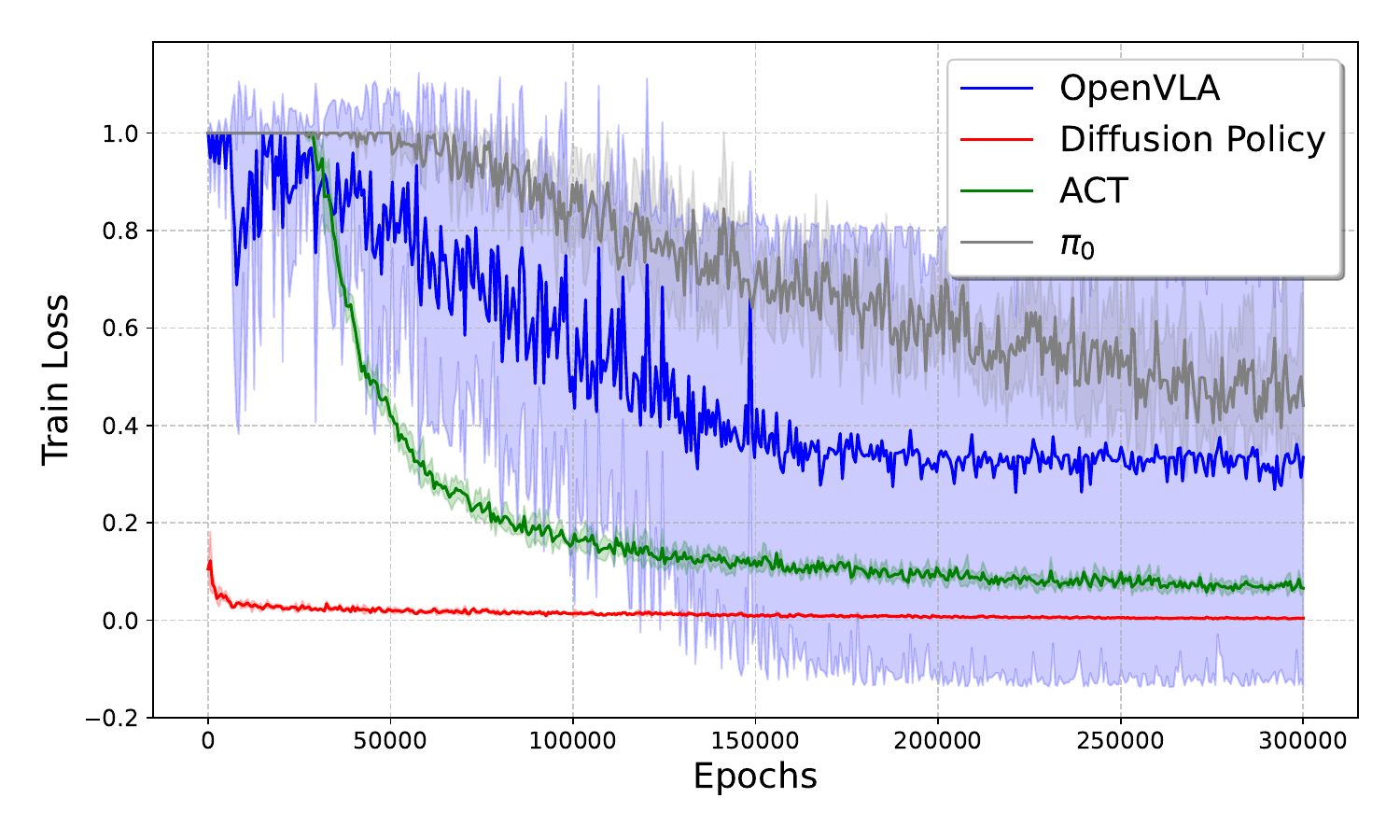}

    \caption{Partial fine-tuning results for VLAs (OpenVLA and $\pi_0$) compared with training Diffusion Policy (DP) and ACT from scratch on our dataset. VLAs require more epochs to converge and show higher performance variance.}

    \label{fig5}
\end{figure}

% Figure 6 ----------
\begin{figure}[t]
  \includegraphics[width=\columnwidth]{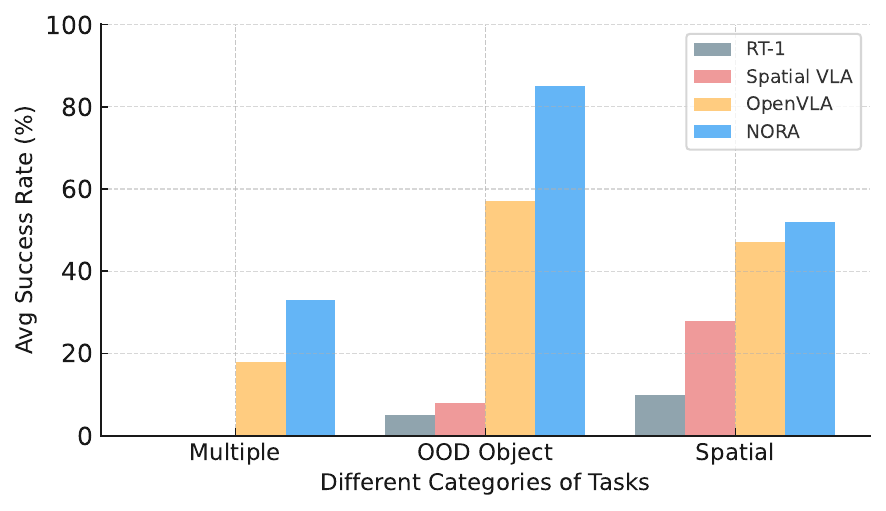}
  \caption {Success rates of fully fine-tuned VLAs on multi-object pick-and-place, out-of-distribution object manipulation, and spatial relationship reasoning tasks. NORA achieves the highest performance.}
  \label{fig:6}
  \vspace{-10 pt}
\end{figure}

\begin{itemize}[leftmargin=1em, itemsep=0pt, topsep=0pt]

    \item \textit{Partial Fine-tuning.} We observe that DP and ACT converge stably with low training variance (Fig.~\ref{fig5}). In contrast, generalist models such as OpenVLA and $\pi_0$ require far more iterations to reach comparable accuracy and exhibit greater variance. This difference reflects model scale: compact task-specific policies like DP and ACT contain only tens of millions of parameters, whereas generalist policies such as OpenVLA (7B) and $\pi_0$ (3B) rely on billion-scale backones, making optimization slower and less stable. Notably, while DP attains lower loss by fitting directly to noise, it still requires additional training to produce coherent action sequences even after loss convergence.

    \item \textit{Full Fine-tuning.} 
    The fine-tuned VLAs are evaluated on three tasks: (1) out-of-distribution object manipulation, (2) spatial relationship reasoning, and (3) multi-object pick-and-place. In task (1), both NORA and OpenVLA succeed, whereas SpatialVLA fails due to incorrect affordance-point estimation. In task (2), NORA follows instructions correctly, while OpenVLA fails and SpatialVLA shows unstable performance. In task (3), only NORA executes the task successfully, while the other models fail to complete it reliably (Fig.~\ref{fig:6}).

\end{itemize}

%\paragraph{Impact of Action Chunking}
%We conducted ablation studies to evaluate the effect of action chunking in both simulation and real-world environments. Results indicate that SpatialVLA and NORA tend to fail more often with chunking enabled, frequently due to collision-induced task failures. However, under high-frequency simulation settings, NORA benefits from action chunking, as shown in Table~\ref{tab:4}, suggesting its effectiveness increases with higher control frequency.

\paragraph{Simulated Skill Adaptation.}
We compare model performance on simulation benchmarks and real robot deployments (Table~\ref{tab:4}). The results show that the finetuned $\pi_0$ model outperforms all baselines across tasks, achieving the highest success rates on Spatial and Object tasks while maintaining strong performance on Goal and Long-horizon tasks. Furthermore, the ablation study of action chunking (AC) on NORA demonstrates that AC consistently enhances performance across most simulation tasks. Notably, a significant drop in performance is observed when transferring from simulation to the real world (Appendix Table~\ref{tab:task_performance}). %The simulation benchmark includes 30 procedurally-generated, disentangled tasks requiring nuanced spatial reasoning (\textit{LIBERO-Spatial}), object understanding (\textit{LIBERO-Object}), and goal interpretation (\textit{LIBERO-Goal}), as well as 10 long-horizon, entangled tasks (\textit{LIBERO-Long}).

\paragraph{Robustness to Perturbations.}

We assess robustness by introducing distractor objects into the environment. As shown in Appendix Table~\ref{tab:distractors}, both OpenVLA and NORA degrade substantially under these perturbations, underscoring their sensitivity to novel conditions.

\paragraph{Key Takeaways.}
Current VLAs still face significant limitations in the following areas:

\begin{itemize}[leftmargin=1em, itemsep=0pt, topsep=0pt]
    \item \textit{Adaptation and Generalization.}  A generic robotic policy is expected to adapt rapidly to datasets with distributional shifts. However, our partial fine-tuning results show that, given their large model capacities and the limited size of task-specific datasets, current VLAs fail to achieve efficient adaptation. While fine-tuning improves performance, it demands extensive data and prolonged training, which is usually impractical for many real-world scenarios.
    
    \item \textit{Robustness.} Robustness to distribution shifts without finetuning remains a critical challenge. %We evaluated model performance under perturbations and compared outcomes in simulation versus real-world settings. 
    Results reveal substantial degradation when encountering unseen objects and during sim-to-real transfer, underscoring the fragility of current VLA models in dynamic and unpredictable environments.
\end{itemize}

These findings indicate that although VLAs hold significant promise, they remain constrained by poor data efficiency, slow adaptation, and brittle robustness. Bridging these gaps will require both algorithmic advances—such as more parameter-efficient adaptation, bias-resistant pretraining, and stronger perception backbones—and system-level improvements in data collection and augmentation. Addressing these limitations is crucial if VLAs are to evolve from research prototypes into reliable, deployable policies for real-world robots operating under uncertainty.

% Discussions ====================================================
\section{Constraints and Future Directions}
\label{sec:5}
Despite the promise of foundation models (FMs) for enabling embodied agents to perform daily tasks, several critical constraints still hinder their reliable deployment:

\paragraph{Data Scarcity.}
Unlike natural language data that abundantly available on the internet, robotic datasets are costly to collect due to hardware wear, safety risks, and labor-intensive demonstrations. A key direction is improving data efficiency through parameter-efficient adaptation, imitation from unlabeled interaction, and self-supervised pretraining. Complementary approaches include leveraging high-fidelity simulators and developing robust sim-to-real transfer pipelines to reduce reliance on large real-world collections.

%\paragraph{Generalization.}
%FM generalize poorly to tasks that are underrepresented in training data, especially for VLAs that depend on datasets with limited scale and diversity in scene layouts and robot embodiments. In addition, as shown in our case study, most models still struggle with fine-grained spatial reasoning and temporal understanding, limiting their ability to align language with complex visual scenes and dynamic interactions.

\paragraph{Efficient Inference.}
VLAs place heavy computational burdens on robotic platforms, leading to bottlenecks in inference speed. The constraints motivate research on lightweight architectures and efficient decoding strategies that can sustain performance while satisfying the real-time requirements of embodied control.

\paragraph{Explainability and Safety.}
Most FMs lack explicit mechanisms for interpretability or safety guarantees—factors that are crucial for deployment in high-stakes or unstructured environments. These models may output confident but incorrect actions when faced with out-of-distribution inputs or adversarial perturbations. Moreover, without built-in constraints to enforce ethical and operational boundaries, VLAs risk misinterpreting ambiguous instructions in ways that compromise human intent or physical safety.

\vspace{20pt}
\section*{Limitations}
For VLA generalization study, this work focuses primarily on task-level performance and does not extensively examine generalization across diverse robot morphologies. Yet this remains a critical challenge for real-world deployment: robots with different embodiments—such as bimanual manipulators, humanoid robots, or mobile platforms—require distinct control protocols and safety constraints. The absence of a generic policy that adapts seamlessly across such morphologies limits the universality of current VLA approaches.

This work also does not directly investigate the grounding of instructions that are open-ended or ambiguous. Existing VLAs are trained largely on curated datasets that map well-structured commands to specific actions, but this reliance constrains their semantic understanding. Consequently, when faced with vague or out-of-distribution instructions, they often fail to infer reasonable behaviors. Addressing this limitation will require integrating richer language-understanding modules and more diverse training data to improve robustness in handling underspecified or ambiguous input.

\section*{Acknowledgment}
This work is supported by the National Robotics Programme (Award \#M23NKBK0091, \#M25N4N2009). This work is also supported by the National Research Foundation, Singapore under its National Large Language Models Funding Initiative (AISG Award No: AISG-NMLP-2024-005), NTU SUG project \#025628-00001:Post-training to Improve Embodied AI Agents, and A*STAR SERC CRF funding to Cheston Tan. We thank Wei Qi Toh and Ray Prithviraj for their assistance with robot experiments, and Shaohua Li for manuscript polishing.

\bibliography{Ref}

\begin{thebibliography}{37}
\providecommand{\natexlab}[1]{#1}

\bibitem[{Bai et~al.(2025)Bai, Chen, Liu, Wang, Ge, Song, Dang, Wang, Wang, Tang, Zhong, Zhu, Yang, Li, Wan, Wang, Ding, Fu, Xu, Ye, Zhang, Xie, Cheng, Zhang, Yang, Xu, and Lin}]{Qwen}
Shuai Bai, Keqin Chen, Xuejing Liu, Jialin Wang, Wenbin Ge, Sibo Song, Kai Dang, Peng Wang, Shijie Wang, Jun Tang, Humen Zhong, Yuanzhi Zhu, Mingkun Yang, Zhaohai Li, Jianqiang Wan, Pengfei Wang, Wei Ding, Zheren Fu, Yiheng Xu, and 8 others. 2025.
\newblock \href {https://doi.org/10.48550/arXiv.2502.13923} {Qwen2.5-{{VL Technical Report}}}.
\newblock \emph{Preprint}, arXiv:2502.13923.

\bibitem[{Bandyopadhyay et~al.(2024)Bandyopadhyay, Talbot et~al.}]{RSS24demo}
Tirthankar Bandyopadhyay, Fletcher Talbot, and 1 others. 2024.
\newblock Demonstrating {{Event-Triggered Investigation}} and {{Sample Collection}} for {{Human Scientists}} using {{Field Robots}} and {{Large Foundation Models}}.
\newblock In \emph{Robotics: {{Science}} and {{Systems}}}.

\bibitem[{Bjorck et~al.(2025)Bjorck, Casta{\~n}eda, Cherniadev, Da, Ding, Fan, Fang, Fox, Hu, Huang et~al.}]{bjorck2025gr00t}
Johan Bjorck, Fernando Casta{\~n}eda, Nikita Cherniadev, Xingye Da, Runyu Ding, Linxi Fan, Yu~Fang, Dieter Fox, Fengyuan Hu, Spencer Huang, and 1 others. 2025.
\newblock Gr00t n1: An open foundation model for generalist humanoid robots.
\newblock \emph{arXiv preprint arXiv:2503.14734}.

\bibitem[{Black et~al.(2024)Black, Brown, Driess, Esmail, Equi, Finn, Fusai, Groom, Hausman, Ichter, Jakubczak, Jones, Ke, Levine, {Li-Bell}, Mothukuri, Nair, Pertsch, Shi, Tanner, Vuong, Walling, Wang, and Zhilinsky}]{black2024pi_0}
Kevin Black, Noah Brown, Danny Driess, Adnan Esmail, Michael Equi, Chelsea Finn, Niccolo Fusai, Lachy Groom, Karol Hausman, Brian Ichter, Szymon Jakubczak, Tim Jones, Liyiming Ke, Sergey Levine, Adrian {Li-Bell}, Mohith Mothukuri, Suraj Nair, Karl Pertsch, Lucy~Xiaoyang Shi, and 5 others. 2024.
\newblock $\pi_0$: A vision-language-action flow model for general robot control.
\newblock \emph{arXiv:2410.24164}.

\bibitem[{Brohan et~al.(2023)Brohan, Brown, Carbajal, Chebotar, Dabis, Finn, Gopalakrishnan, Hausman, Herzog, Hsu, Ibarz, Ichter, Irpan, Jackson, Jesmonth, Joshi, Julian, Kalashnikov, Kuang, Leal, Lee, Levine, Lu, Malla, Manjunath, Mordatch, Nachum, Parada, Peralta, Perez, Pertsch, Quiambao, Rao, Ryoo, Salazar, Sanketi, Sayed, Singh, Sontakke, Stone, Tan, Tran, Vanhoucke, Vega, Vuong, Xia, Xiao, Xu, Xu, Yu, and Zitkovich}]{brohan2022rt}
Anthony Brohan, Noah Brown, Justice Carbajal, Yevgen Chebotar, Joseph Dabis, Chelsea Finn, Keerthana Gopalakrishnan, Karol Hausman, Alex Herzog, Jasmine Hsu, Julian Ibarz, Brian Ichter, Alex Irpan, Tomas Jackson, Sally Jesmonth, Nikhil~J Joshi, Ryan Julian, Dmitry Kalashnikov, Yuheng Kuang, and 32 others. 2023.
\newblock Rt-1: Robotics transformer for real-world control at scale.
\newblock In \emph{Proceedings of Robotics: Science and Systems}.

\bibitem[{Bu et~al.(2025)Bu, Yang, Cai, Gao, Ren, Yao, Luo, and Li}]{bulearning}
Qingwen Bu, Yanting Yang, Jisong Cai, Shenyuan Gao, Guanghui Ren, Maoqing Yao, Ping Luo, and Hongyang Li. 2025.
\newblock Learning to act anywhere with task-centric latent actions.
\newblock \emph{arXiv preprint arXiv:2502.14420}.

\bibitem[{Chi et~al.(2023)Chi, Xu, Feng, Cousineau, Du, Burchfiel, Tedrake, and Song}]{chi2023diffusion}
Cheng Chi, Zhenjia Xu, Siyuan Feng, Eric Cousineau, Yilun Du, Benjamin Burchfiel, Russ Tedrake, and Shuran Song. 2023.
\newblock Diffusion policy: Visuomotor policy learning via action diffusion.
\newblock \emph{The International Journal of Robotics Research}.

\bibitem[{Collaboration et~al.(2024)Collaboration, O'Neill, Rehman, Gupta, Maddukuri, Gupta, Padalkar, Lee, Pooley, Gupta, Mandlekar, Jain, Tung, Bewley, Herzog, Irpan, Khazatsky, Rai, Gupta, Wang, Kolobov, Singh, Garg, Kembhavi, Xie, Brohan, Raffin, Sharma, Yavary, Jain, Balakrishna, Wahid, Burgess-Limerick, Kim, Schölkopf, Wulfe, Ichter, Lu, Xu, Le, Finn, Wang, Xu, Chi, Huang, Chan, Agia, Pan, Fu, Devin, Xu, Morton, Driess, Chen, Pathak, Shah, Büchler, Jayaraman, Kalashnikov, Sadigh, Johns, Foster, Liu, Ceola, Xia, Zhao, Frujeri, Stulp, Zhou, Sukhatme, Salhotra, Yan, Feng, Schiavi, Berseth, Kahn, Yang, Wang, Su, Fang, Shi, Bao, Amor, Christensen, Furuta, Bharadhwaj, Walke, Fang, Ha, Mordatch, Radosavovic, Leal, Liang, Abou-Chakra, Kim, Drake, Peters, Schneider, Hsu, Vakil, Bohg, Bingham, Wu, Gao, Hu, Wu, Wu, Sun, Luo, Gu, Tan, Oh, Wu, Lu, Yang, Malik, Silvério, Hejna, Booher, Tompson, Yang, Salvador, Lim, Han, Wang, Rao, Pertsch, Hausman, Go, Gopalakrishnan, Goldberg, Byrne, Oslund, Kawaharazuka, Black,
  Lin, Zhang, Ehsani, Lekkala, Ellis, Rana, Srinivasan, Fang, Singh, Zeng, Hatch, Hsu, Itti, Chen, Pinto, Fei-Fei, Tan, Fan, Ott, Lee, Weihs, Chen, Lepert, Memmel, Tomizuka, Itkina, Castro, Spero, Du, Ahn, Yip, Zhang, Ding, Heo, Srirama, Sharma, Kim, Irshad, Kanazawa, Hansen, Heess, Joshi, Suenderhauf, Liu, Palo, Shafiullah, Mees, Kroemer, Bastani, Sanketi, Miller, Yin, Wohlhart, Xu, Fagan, Mitrano, Sermanet, Abbeel, Sundaresan, Chen, Vuong, Rafailov, Tian, Doshi, Martín-Martín, Baijal, Scalise, Hendrix, Lin, Qian, Zhang, Mendonca, Shah, Hoque, Julian, Bustamante, Kirmani, Levine, Lin, Moore, Bahl, Dass, Sonawani, Tulsiani, Song, Xu, Haldar, Karamcheti, Adebola, Guist, Nasiriany, Schaal, Welker, Tian, Ramamoorthy, Dasari, Belkhale, Park, Nair, Mirchandani, Osa, Gupta, Harada, Matsushima, Xiao, Kollar, Yu, Ding, Davchev, Zhao, Armstrong, Darrell, Chung, Jain, Kumar, Vanhoucke, Guizilini, Zhan, Zhou, Burgard, Chen, Chen, Wang, Zhu, Geng, Liu, Liangwei, Li, Pang, Lu, Ma, Kim, Chebotar, Zhou, Zhu, Wu, Xu, Wang,
  Bisk, Dou, Cho, Lee, Cui, Cao, Wu, Tang, Zhu, Zhang, Jiang, Li, Li, Iwasawa, Matsuo, Ma, Xu, Cui, Zhang, Fu, and Lin}]{o2024open}
Embodiment Collaboration, Abby O'Neill, Abdul Rehman, Abhinav Gupta, Abhiram Maddukuri, Abhishek Gupta, Abhishek Padalkar, Abraham Lee, Acorn Pooley, Agrim Gupta, Ajay Mandlekar, Ajinkya Jain, Albert Tung, Alex Bewley, Alex Herzog, Alex Irpan, Alexander Khazatsky, Anant Rai, Anchit Gupta, and 275 others. 2024.
\newblock Open x-embodiment: Robotic learning datasets and rt-x models.
\newblock In \emph{ICRA}.

\bibitem[{Fan et~al.(2025)Fan, Yang, Liu, Wu, Che, Liu, and Wan}]{fan2025diffusion}
Shichao Fan, Quantao Yang, Yajie Liu, Kun Wu, Zhengping Che, Qingjie Liu, and Min Wan. 2025.
\newblock Diffusion trajectory-guided policy for long-horizon robot manipulation.
\newblock \emph{arXiv preprint arXiv:2502.10040}.

\bibitem[{Gemini Robotics~Team(2025)}]{gemini}
Google~DeepMind Gemini Robotics~Team. 2025.
\newblock \href {https://doi.org/10.48550/arXiv.2503.20020} {Gemini {{Robotics}}: {{Bringing AI}} into the {{Physical World}}}.
\newblock \emph{Preprint}, arXiv:2503.20020.

\bibitem[{Glocker et~al.(2025)Glocker, H{\"o}nig, Hirschmanner, and Vincze}]{hub_llama3.2v}
Marc Glocker, Peter H{\"o}nig, Matthias Hirschmanner, and Markus Vincze. 2025.
\newblock \href {https://doi.org/10.48550/arXiv.2504.21716} {{{LLM-Empowered Embodied Agent}} for {{Memory-Augmented Task Planning}} in {{Household Robotics}}}.
\newblock \emph{Preprint}, arXiv:2504.21716.

\bibitem[{{Google DeepMind}(2024)}]{deepmind2024gemini}
{Google DeepMind}. 2024.
\newblock \href {https://blog.google/technology/google-deepmind/google-gemini-ai-update-december-2024/#ceo-message} {Gemini: Our largest and most capable ai models yet}.
\newblock Accessed: 2025-05-17.

\bibitem[{Grattafiori et~al.(2024)Grattafiori, Dubey, Jauhri, Pandey, Kadian, Al-Dahle, Letman, Mathur, Schelten, Vaughan, Yang, Fan, Goyal, Hartshorn, Yang, Mitra, Sravankumar, Korenev, Hinsvark, Rao, Zhang, Rodriguez, Gregerson, Spataru, Roziere, Biron, Tang, Chern, Caucheteux, Nayak, Bi, Marra, McConnell, Keller, Touret, Wu, Wong, Ferrer, Nikolaidis, Allonsius, Song, Pintz, Livshits, Wyatt, Esiobu, Choudhary, Mahajan, Garcia-Olano, Perino, Hupkes, Lakomkin, AlBadawy, Lobanova, Dinan, Smith, Radenovic, Guzmán, Zhang, Synnaeve, Lee, Anderson, Thattai, Nail, Mialon, Pang, Cucurell, Nguyen, Korevaar, Xu, Touvron, Zarov, Ibarra, Kloumann, Misra, Evtimov, Zhang, Copet, Lee, Geffert, Vranes, Park, Mahadeokar, Shah, van~der Linde, Billock, Hong, Lee, Fu, Chi, Huang, Liu, Wang, Yu, Bitton, Spisak, Park, Rocca, Johnstun, Saxe, Jia, Alwala, Prasad, Upasani, Plawiak, Li, Heafield, Stone, El-Arini, Iyer, Malik, Chiu, Bhalla, Lakhotia, Rantala-Yeary, van~der Maaten, Chen, Tan, Jenkins, Martin, Madaan, Malo, Blecher,
  Landzaat, de~Oliveira, Muzzi, Pasupuleti, Singh, Paluri, Kardas, Tsimpoukelli, Oldham, Rita, Pavlova, Kambadur, Lewis, Si, Singh, Hassan, Goyal, Torabi, Bashlykov, Bogoychev, Chatterji, Zhang, Duchenne, Çelebi, Alrassy, Zhang, Li, Vasic, Weng, Bhargava, Dubal, Krishnan, Koura, Xu, He, Dong, Srinivasan, Ganapathy, Calderer, Cabral, Stojnic, Raileanu, Maheswari, Girdhar, Patel, Sauvestre, Polidoro, Sumbaly, Taylor, Silva, Hou, Wang, Hosseini, Chennabasappa, Singh, Bell, Kim, Edunov, Nie, Narang, Raparthy, Shen, Wan, Bhosale, Zhang, Vandenhende, Batra, Whitman, Sootla, Collot, Gururangan, Borodinsky, Herman, Fowler, Sheasha, Georgiou, Scialom, Speckbacher, Mihaylov, Xiao, Karn, Goswami, Gupta, Ramanathan, Kerkez, Gonguet, Do, Vogeti, Albiero, Petrovic, Chu, Xiong, Fu, Meers, Martinet, Wang, Wang, Tan, Xia, Xie, Jia, Wang, Goldschlag, Gaur, Babaei, Wen, Song, Zhang, Li, Mao, Coudert, Yan, Chen, Papakipos, Singh, Srivastava, Jain, Kelsey, Shajnfeld, Gangidi, Victoria, Goldstand, Menon, Sharma, Boesenberg,
  Baevski, Feinstein, Kallet, Sangani, Teo, Yunus, Lupu, Alvarado, Caples, Gu, Ho, Poulton, Ryan, Ramchandani, Dong, Franco, Goyal, Saraf, Chowdhury, Gabriel, Bharambe, Eisenman, Yazdan, James, Maurer, Leonhardi, Huang, Loyd, Paola, Paranjape, Liu, Wu, Ni, Hancock, Wasti, Spence, Stojkovic, Gamido, Montalvo, Parker, Burton, Mejia, Liu, Wang, Kim, Zhou, Hu, Chu, Cai, Tindal, Feichtenhofer, Gao, Civin, Beaty, Kreymer, Li, Adkins, Xu, Testuggine, David, Parikh, Liskovich, Foss, Wang, Le, Holland, Dowling, Jamil, Montgomery, Presani, Hahn, Wood, Le, Brinkman, Arcaute, Dunbar, Smothers, Sun, Kreuk, Tian, Kokkinos, Ozgenel, Caggioni, Kanayet, Seide, Florez, Schwarz, Badeer, Swee, Halpern, Herman, Sizov, Guangyi, Zhang, Lakshminarayanan, Inan, Shojanazeri, Zou, Wang, Zha, Habeeb, Rudolph, Suk, Aspegren, Goldman, Zhan, Damlaj, Molybog, Tufanov, Leontiadis, Veliche, Gat, Weissman, Geboski, Kohli, Lam, Asher, Gaya, Marcus, Tang, Chan, Zhen, Reizenstein, Teboul, Zhong, Jin, Yang, Cummings, Carvill, Shepard, McPhie,
  Torres, Ginsburg, Wang, Wu, U, Saxena, Khandelwal, Zand, Matosich, Veeraraghavan, Michelena, Li, Jagadeesh, Huang, Chawla, Huang, Chen, Garg, A, Silva, Bell, Zhang, Guo, Yu, Moshkovich, Wehrstedt, Khabsa, Avalani, Bhatt, Mankus, Hasson, Lennie, Reso, Groshev, Naumov, Lathi, Keneally, Liu, Seltzer, Valko, Restrepo, Patel, Vyatskov, Samvelyan, Clark, Macey, Wang, Hermoso, Metanat, Rastegari, Bansal, Santhanam, Parks, White, Bawa, Singhal, Egebo, Usunier, Mehta, Laptev, Dong, Cheng, Chernoguz, Hart, Salpekar, Kalinli, Kent, Parekh, Saab, Balaji, Rittner, Bontrager, Roux, Dollar, Zvyagina, Ratanchandani, Yuvraj, Liang, Alao, Rodriguez, Ayub, Murthy, Nayani, Mitra, Parthasarathy, Li, Hogan, Battey, Wang, Howes, Rinott, Mehta, Siby, Bondu, Datta, Chugh, Hunt, Dhillon, Sidorov, Pan, Mahajan, Verma, Yamamoto, Ramaswamy, Lindsay, Lindsay, Feng, Lin, Zha, Patil, Shankar, Zhang, Zhang, Wang, Agarwal, Sajuyigbe, Chintala, Max, Chen, Kehoe, Satterfield, Govindaprasad, Gupta, Deng, Cho, Virk, Subramanian, Choudhury,
  Goldman, Remez, Glaser, Best, Koehler, Robinson, Li, Zhang, Matthews, Chou, Shaked, Vontimitta, Ajayi, Montanez, Mohan, Kumar, Mangla, Ionescu, Poenaru, Mihailescu, Ivanov, Li, Wang, Jiang, Bouaziz, Constable, Tang, Wu, Wang, Wu, Gao, Kleinman, Chen, Hu, Jia, Qi, Li, Zhang, Zhang, Adi, Nam, Yu, Wang, Zhao, Hao, Qian, Li, He, Rait, DeVito, Rosnbrick, Wen, Yang, Zhao, and Ma}]{llamaVL}
Aaron Grattafiori, Abhimanyu Dubey, Abhinav Jauhri, Abhinav Pandey, Abhishek Kadian, Ahmad Al-Dahle, Aiesha Letman, Akhil Mathur, Alan Schelten, Alex Vaughan, Amy Yang, Angela Fan, Anirudh Goyal, Anthony Hartshorn, Aobo Yang, Archi Mitra, Archie Sravankumar, Artem Korenev, Arthur Hinsvark, and 542 others. 2024.
\newblock \href {https://doi.org/10.48550/arXiv.2407.21783} {The {{Llama}} 3 {{Herd}} of {{Models}}}.

\bibitem[{Hung et~al.(2025)Hung, Sun, Hong, Zadeh, Li, Tan, Majumder, and Poria}]{hung2025nora}
Chia-Yu Hung, Qi~Sun, Pengfei Hong, Amir Zadeh, Chuan Li, U-Xuan Tan, Navonil Majumder, and Soujanya Poria. 2025.
\newblock Nora: A small open-sourced generalist vision language action model for embodied tasks.
\newblock \emph{arXiv preprint arXiv:2504.19854}.

\bibitem[{Intelligence et~al.(2025)Intelligence, Black, Brown, Darpinian, Dhabalia, Driess, Esmail, Equi, Finn, Fusai, Galliker, Ghosh, Groom, Hausman, Ichter, Jakubczak, Jones, Ke, LeBlanc, Levine, Li-Bell, Mothukuri, Nair, Pertsch, Ren, Shi, Smith, Springenberg, Stachowicz, Tanner, Vuong, Walke, Walling, Wang, Yu, and Zhilinsky}]{intelligence2025pi_}
Physical Intelligence, Kevin Black, Noah Brown, James Darpinian, Karan Dhabalia, Danny Driess, Adnan Esmail, Michael Equi, Chelsea Finn, Niccolo Fusai, Manuel~Y. Galliker, Dibya Ghosh, Lachy Groom, Karol Hausman, Brian Ichter, Szymon Jakubczak, Tim Jones, Liyiming Ke, Devin LeBlanc, and 17 others. 2025.
\newblock $\pi_{0.5}$: a vision-language-action model with open-world generalization.
\newblock \emph{arXiv preprint arXiv:2504.16054}.

\bibitem[{Kim et~al.(2024)Kim, Pertsch, Karamcheti, Xiao, Balakrishna, Nair, Rafailov, Foster, Lam, Sanketi, Vuong, Kollar, Burchfiel, Tedrake, Sadigh, Levine, Liang, and Finn}]{kim2024openvla}
Moo~Jin Kim, Karl Pertsch, Siddharth Karamcheti, Ted Xiao, Ashwin Balakrishna, Suraj Nair, Rafael Rafailov, Ethan Foster, Grace Lam, Pannag Sanketi, Quan Vuong, Thomas Kollar, Benjamin Burchfiel, Russ Tedrake, Dorsa Sadigh, Sergey Levine, Percy Liang, and Chelsea Finn. 2024.
\newblock Openvla: An open-source vision-language-action model.
\newblock \emph{arXiv:2406.09246}.

\bibitem[{Li et~al.(2024{\natexlab{a}})Li, Zhu, Xu, Gu, Zhu, Liu, Liu, Peng, Feng, and Tang}]{hub_inhouse}
Jinming Li, Yichen Zhu, Zhiyuan Xu, Jindong Gu, Minjie Zhu, Xin Liu, Ning Liu, Yaxin Peng, Feifei Feng, and Jian Tang. 2024{\natexlab{a}}.
\newblock \href {https://doi.org/10.48550/arXiv.2406.19693} {{{MMRo}}: {{Are Multimodal LLMs Eligible}} as the {{Brain}} for {{In-Home Robotics}}?}

\bibitem[{Li et~al.(2023)Li, Li, Savarese, and Hoi}]{blip}
Junnan Li, Dongxu Li, Silvio Savarese, and Steven Hoi. 2023.
\newblock Blip-2: bootstrapping language-image pre-training with frozen image encoders and large language models.
\newblock In \emph{Proceedings of the 40th International Conference on Machine Learning}, ICML'23. JMLR.org.

\bibitem[{Li et~al.(2024{\natexlab{b}})Li, Liang, Wang, Luo, Chen, Liao, Wei, Deng, Xu, Zhang, Wang, Liu, Fu, Bao, Chen, Shi, Yang, and Guo}]{li2024cogact}
Qixiu Li, Yaobo Liang, Zeyu Wang, Lin Luo, Xi~Chen, Mozheng Liao, Fangyun Wei, Yu~Deng, Sicheng Xu, Yizhong Zhang, Xiaofan Wang, Bei Liu, Jianlong Fu, Jianmin Bao, Dong Chen, Yuanchun Shi, Jiaolong Yang, and Baining Guo. 2024{\natexlab{b}}.
\newblock Cogact: A foundational vision-language-action model for synergizing cognition and action in robotic manipulation.
\newblock \emph{arXiv preprint arXiv:2411.19650}.

\bibitem[{Li et~al.(2024{\natexlab{c}})Li, Zhang, Geng, Geng, Long, Shen, Zhang, Liu, and Dong}]{maniLLM}
Xiaoqi Li, Mingxu Zhang, Yiran Geng, Haoran Geng, Yuxing Long, Yan Shen, Renrui Zhang, Jiaming Liu, and Hao Dong. 2024{\natexlab{c}}.
\newblock Manipllm: Embodied multimodal large language model for object-centric robotic manipulation.
\newblock In \emph{CVPR}.

\bibitem[{Liu et~al.(2023)Liu, Zhu, Gao, Feng, Liu, Zhu, and Stone}]{liu2023libero}
Bo~Liu, Yifeng Zhu, Chongkai Gao, Yihao Feng, Qiang Liu, Yuke Zhu, and Peter Stone. 2023.
\newblock Libero: Benchmarking knowledge transfer for lifelong robot learning.
\newblock \emph{Advances in Neural Information Processing Systems}, 36:44776--44791.

\bibitem[{Liu et~al.(2024{\natexlab{a}})Liu, Zeng, Ren, Li, Zhang, Yang, Jiang, Li, Yang, Su, Zhu, and Zhang}]{groundingdino}
Shilong Liu, Zhaoyang Zeng, Tianhe Ren, Feng Li, Hao Zhang, Jie Yang, Qing Jiang, Chunyuan Li, Jianwei Yang, Hang Su, Jun Zhu, and Lei Zhang. 2024{\natexlab{a}}.
\newblock Grounding dino: Marrying dino with grounded pre-training for open-set object detection.
\newblock In \emph{ECCV}.

\bibitem[{Liu et~al.(2024{\natexlab{b}})Liu, Wu, Li, Tan, Chen, Wang, Xu, Su, and Zhu}]{liu2024rdt}
Songming Liu, Lingxuan Wu, Bangguo Li, Hengkai Tan, Huayu Chen, Zhengyi Wang, Ke~Xu, Hang Su, and Jun Zhu. 2024{\natexlab{b}}.
\newblock Rdt-1b: a diffusion foundation model for bimanual manipulation.
\newblock \emph{arXiv:2410.07864}.

\bibitem[{Lu et~al.(2024)Lu, Liu, Zhang, Wang, Dong, Liu, Sun, Ren, Li, Yang, Sun, Deng, Xu, Xie, and Ruan}]{DS}
Haoyu Lu, Wen Liu, Bo~Zhang, Bingxuan Wang, Kai Dong, Bo~Liu, Jingxiang Sun, Tongzheng Ren, Zhuoshu Li, Hao Yang, Yaofeng Sun, Chengqi Deng, Hanwei Xu, Zhenda Xie, and Chong Ruan. 2024.
\newblock \href {https://doi.org/10.48550/arXiv.2403.05525} {{{DeepSeek-VL}}: {{Towards Real-World Vision-Language Understanding}}}.
\newblock \emph{Preprint}, arXiv:2403.05525.

\bibitem[{Qu et~al.(2025)Qu, Song, Chen, Yao, Ye, Ding, Wang, Gu, Zhao, Wang, and Li}]{qu2025spatialvla}
Delin Qu, Haoming Song, Qizhi Chen, Yuanqi Yao, Xinyi Ye, Yan Ding, Zhigang Wang, JiaYuan Gu, Bin Zhao, Dong Wang, and Xuelong Li. 2025.
\newblock Spatialvla: Exploring spatial representations for visual-language-action model.
\newblock \emph{arXiv preprint arXiv:2501.15830}.

\bibitem[{Ravi et~al.(2025)Ravi, Gabeur, Hu, Hu, Ryali, Ma, Khedr, Rädle, Rolland, Gustafson, Mintun, Pan, Alwala, Carion, Wu, Girshick, Dollár, and Feichtenhofer}]{ravi2024sam}
Nikhila Ravi, Valentin Gabeur, Yuan-Ting Hu, Ronghang Hu, Chaitanya Ryali, Tengyu Ma, Haitham Khedr, Roman Rädle, Chloe Rolland, Laura Gustafson, Eric Mintun, Junting Pan, Kalyan~Vasudev Alwala, Nicolas Carion, Chao-Yuan Wu, Ross Girshick, Piotr Dollár, and Christoph Feichtenhofer. 2025.
\newblock \href {https://openreview.net/forum?id=Ha6RTeWMd0} {{SAM} 2: Segment anything in images and videos}.
\newblock In \emph{ICLR}.

\bibitem[{Ren et~al.(2024)Ren, Liu, Zeng, Lin, Li, Cao, Chen, Huang, Chen, Yan, Zeng, Zhang, Li, Yang, Li, Jiang, and Zhang}]{ren2024grounded}
Tianhe Ren, Shilong Liu, Ailing Zeng, Jing Lin, Kunchang Li, He~Cao, Jiayu Chen, Xinyu Huang, Yukang Chen, Feng Yan, Zhaoyang Zeng, Hao Zhang, Feng Li, Jie Yang, Hongyang Li, Qing Jiang, and Lei Zhang. 2024.
\newblock Grounded sam: Assembling open-world models for diverse visual tasks.
\newblock \emph{arXiv:2401.14159}.

\bibitem[{Reuss et~al.(2024)Reuss, Ya{\u{g}}murlu, Wenzel, and Lioutikov}]{reuss2024multimodal}
Moritz Reuss, {\"O}mer~Erdin{\c{c}} Ya{\u{g}}murlu, Fabian Wenzel, and Rudolf Lioutikov. 2024.
\newblock Multimodal diffusion transformer: Learning versatile behavior from multimodal goals.
\newblock In \emph{Robotics: Science and Systems}.

\bibitem[{Sun et~al.(2024)Sun, Hong, Pala, Toh, Tan, Ghosal, and Poria}]{sun2024emma}
Qi~Sun, Pengfei Hong, Tej~Deep Pala, Vernon Toh, U-Xuan Tan, Deepanway Ghosal, and Soujanya Poria. 2024.
\newblock Emma-x: An embodied multimodal action model with grounded chain of thought and look-ahead spatial reasoning.
\newblock \emph{arXiv:2412.11974}.

\bibitem[{Team et~al.(2024)Team, Ghosh, Walke, Pertsch, Black, Mees, Dasari, Hejna, Kreiman, Xu, Luo, Tan, Chen, Sanketi, Vuong, Xiao, Sadigh, Finn, and Levine}]{team2024octo}
Octo~Model Team, Dibya Ghosh, Homer Walke, Karl Pertsch, Kevin Black, Oier Mees, Sudeep Dasari, Joey Hejna, Tobias Kreiman, Charles Xu, Jianlan Luo, You~Liang Tan, Lawrence~Yunliang Chen, Pannag Sanketi, Quan Vuong, Ted Xiao, Dorsa Sadigh, Chelsea Finn, and Sergey Levine. 2024.
\newblock Octo: An open-source generalist robot policy.
\newblock In \emph{Proceedings of Robotics: Science and Systems}.

\bibitem[{Wen et~al.(2024)Wen, Zhu, Zhu, Tang, Li, Zhou, Li, Liu, Peng, Shen, and Feng}]{wen2024diffusion}
Junjie Wen, Minjie Zhu, Yichen Zhu, Zhibin Tang, Jinming Li, Zhongyi Zhou, Chengmeng Li, Xiaoyu Liu, Yaxin Peng, Chaomin Shen, and Feifei Feng. 2024.
\newblock Diffusion-vla: Scaling robot foundation models via unified diffusion and autoregression.
\newblock \emph{arXiv:2412.03293}.

\bibitem[{Wen et~al.(2025)Wen, Zhu, Li, Tang, Shen, and Feng}]{wen2025dexvla}
Junjie Wen, Yichen Zhu, Jinming Li, Zhibin Tang, Chaomin Shen, and Feifei Feng. 2025.
\newblock Dexvla: Vision-language model with plug-in diffusion expert for general robot control.
\newblock \emph{arXiv preprint arXiv:2502.05855}.

\bibitem[{Werby et~al.(2024)Werby, Huang, B{\"u}chner, Valada, and Burgard}]{3dSceneGraph}
Abdelrhman Werby, Chenguang Huang, Martin B{\"u}chner, Abhinav Valada, and Wolfram Burgard. 2024.
\newblock \href {https://doi.org/10.15607/RSS.2024.XX.077} {Hierarchical {{Open-Vocabulary 3D Scene Graphs}} for {{Language-Grounded Robot Navigation}}}.
\newblock In \emph{Robotics: {{Science}} and {{Systems XX}}}.

\bibitem[{Yang et~al.(2024)Yang, Feng, Chen, Park, Wang, Dou, Zeng, Chen, Gangopadhyay, Owens, and Wong}]{yang2024binding}
Fengyu Yang, Chao Feng, Ziyang Chen, Hyoungseob Park, Daniel Wang, Yiming Dou, Ziyao Zeng, Xien Chen, Rit Gangopadhyay, Andrew Owens, and Alex Wong. 2024.
\newblock Binding touch to everything: Learning unified multimodal tactile representations.
\newblock In \emph{CVPR}, pages 26340--26353.

\bibitem[{Zhao et~al.(2024)Zhao, Ma, Wang, and Adelson}]{zhao2024transferable}
Jialiang Zhao, Yuxiang Ma, Lirui Wang, and Edward~H. Adelson. 2024.
\newblock \href {https://openreview.net/forum?id=KXsropnmNI} {Transferable tactile transformers for representation learning across diverse sensors and tasks}.
\newblock In \emph{CoRL}.

\bibitem[{Zheng et~al.(2025)Zheng, Liang, Huang, Gao, III, Kolobov, Huang, and Yang}]{zheng2024tracevla}
Ruijie Zheng, Yongyuan Liang, Shuaiyi Huang, Jianfeng Gao, Hal~Daum{\'e} III, Andrey Kolobov, Furong Huang, and Jianwei Yang. 2025.
\newblock \href {https://openreview.net/forum?id=b1CVu9l5GO} {Trace{VLA}: Visual trace prompting enhances spatial-temporal awareness for generalist robotic policies}.
\newblock In \emph{ICLR}.

\bibitem[{Zitkovich et~al.(2023)Zitkovich, Yu, Xu, Xu, Xiao, Xia, Wu, Wohlhart, Welker, Wahid, Vuong, Vanhoucke, Tran, Soricut, Singh, Singh, Sermanet, Sanketi, Salazar, Ryoo, Reymann, Rao, Pertsch, Mordatch, Michalewski, Lu, Levine, Lee, Lee, Leal, Kuang, Kalashnikov, Julian, Joshi, Irpan, brian ichter, Hsu, Herzog, Hausman, Gopalakrishnan, Fu, Florence, Finn, Dubey, Driess, Ding, Choromanski, Chen, Chebotar, Carbajal, Brown, Brohan, Arenas, and Han}]{brohan2023rt}
Brianna Zitkovich, Tianhe Yu, Sichun Xu, Peng Xu, Ted Xiao, Fei Xia, Jialin Wu, Paul Wohlhart, Stefan Welker, Ayzaan Wahid, Quan Vuong, Vincent Vanhoucke, Huong Tran, Radu Soricut, Anikait Singh, Jaspiar Singh, Pierre Sermanet, Pannag~R Sanketi, Grecia Salazar, and 35 others. 2023.
\newblock \href {https://openreview.net/forum?id=XMQgwiJ7KSX} {{RT}-2: Vision-language-action models transfer web knowledge to robotic control}.
\newblock In \emph{CoRL}.

\end{thebibliography}
%\bibliographystyle{acl_natbib}

% Appendix =========================================================== 
\clearpage  
\appendix   
\section*{\LARGE Appendix}
\label{sec:appendix}

% PART 1 -------------------
\section{Benchmark Dataset}
\subsection{Cluttered Tabletop Manipulation Dataset} 
\label{app:data}
To evaluate the finetuning behavior of various VLA models under distribution shift, we constructed a custom cluttered tabletop environment using a UR5 robotic arm with a wrist-mounted RealSense RGB-D camera. This setup differs from all existing configurations in the Open-X-Embodiment dataset. Demonstrations for a screwdriver-picking task—amid distractor objects—were collected via teleoperation using a SpaceMouse device. In total, we gathered 163 demonstration episodes\footnote{\url{https://huggingface.co/datasets/bittdy/pick_screw}}. Each episode began with a randomized initial robot pose, followed by an attempt to grasp the target screwdriver.

\subsection{Complex Instruction Grounding Dataset}

We curated an evaluation dataset for the  \textit{complex instruction grounding task} in cluttered scenes\footnote{\url{https://github.com/xiuchao/InstructionGrounding}}. Thirty images were sampled from the action sequences and subsequently categorized based on the number of objects: \textsc{Easy} (<15), \textsc{Medium} (15$\sim$20), and 
 \textsc{Hard} (>20). Objects in the visual scenes were manually annotated using visual prompts and paired with various instructions. The spatial relationship words were illustrate in Table~\ref{tab:A3}.
\begin{table}[h]
    \centering
    \adjustbox{max width=\linewidth}{
    \begin{tabular}{m{1.5cm}m{3cm}m{3cm}}
        \hline
        \textbf{Words} & \textit{Positional}: \hspace{5em} left, right, between, beside, near, far, front, behind & \textit{Directional}: \hspace{3em} aligned with, perpendicular to \\
        \hline
        \textbf{Instruction} & hand over the \text{screwdriver} [on the left of] the red ball. & pass me the screwdriver [aligned with] the marker. \\
        \hline
    \end{tabular}
    }
    \caption{Template words and corresponding examples of generated relation-based instructions for case studies.}
    \label{tab:A3}
\end{table}

For the complex instruction grounding task, including an annotated visual prompt paired with complex instructions in three forms:
implicit, explicit with attributes and spatial references. Additionally, the dataset includes multi-turn questions that refer to more than one object, enabling foundation models to ask clarifying questions to identify the correct object.

\begin{itemize}[leftmargin=1em, itemsep=2pt, topsep=2pt]
    \vspace{5pt}
    \item \textbf{Implicit Instructions.} Here, objects are not explicitly mentioned by name or attributes but are instead described by their functions. This category evaluates the VLMs' ability to infer the correct object based on its use. For example, the dataset includes instructions referring to objects like scissors, screwdrivers, and rulers based on their respective functions.
\begin{figure}[t]
    \centering
    \includegraphics[width=\linewidth]{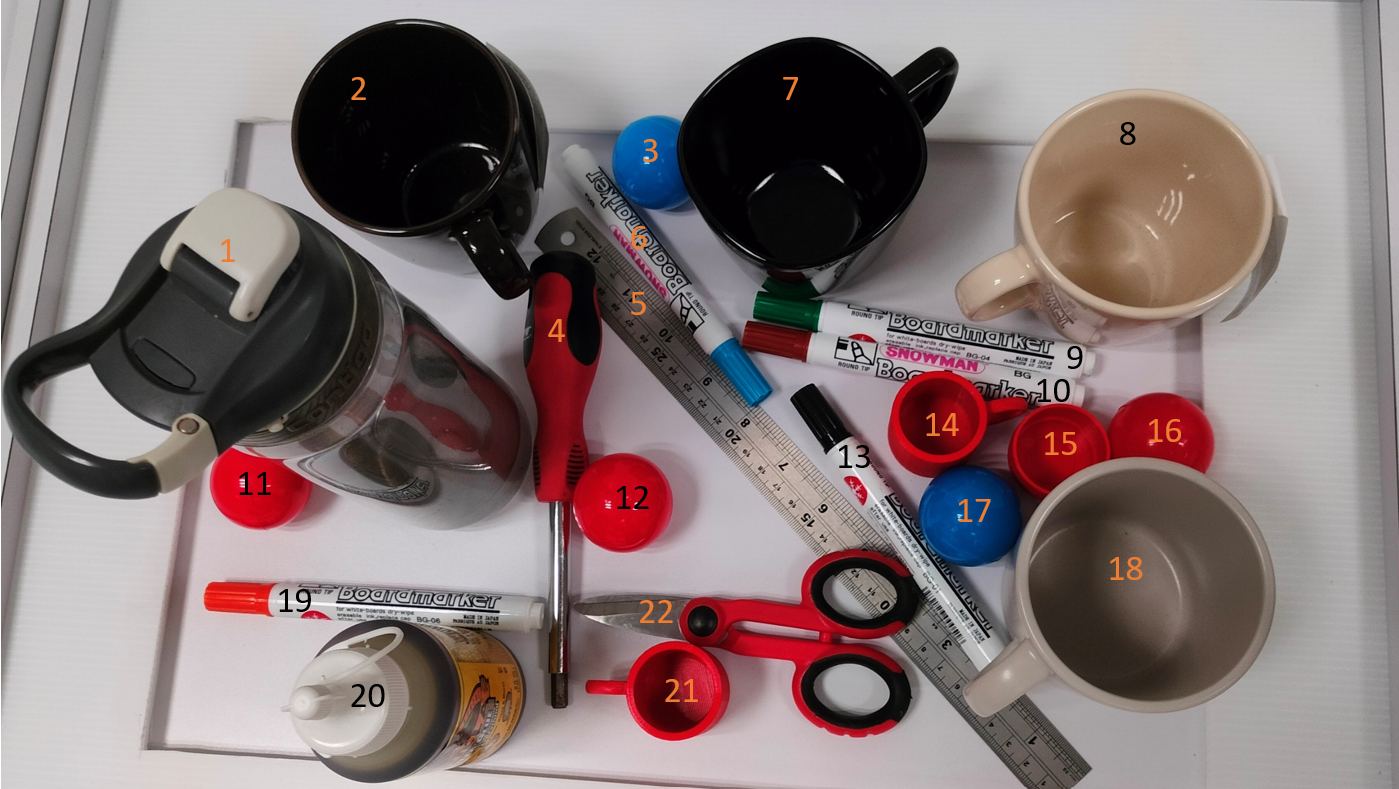}
    \caption{Example of visual prompts}
    \label{fig7}
\end{figure}

    \item \textbf{Explicit Attributes.} In this category, instructions prompt VLMs to identify objects belonging to a category with multiple instances, where each instance can be uniquely identified by explicitly mentioned attributes. In Fig.~\ref{fig7}, the beige mug and the gray mug are included because they are unique when described with attributes. However, objects like the black mug or scissors are excluded. This is because there are two identical black mugs, making them non-unique, and there is only one pair of scissors, which does not require attributes for identification.
    
    \item \textbf{Explicit Relationships.} In this category, instructions describe objects by their spatial relationships to other objects in the image. We ensure that each referenced object is unique within the image. For example, the measuring cup to the right of the screwdriver uniquely identifies the object. These instructions are designed to test the VLMs' ability to comprehend and resolve location-based relationships.

    \item \textbf{Multi-Referent Instructions.} This category includes instructions that correspond to multiple valid objects in the scene. For example, in Fig.~\ref{fig7}, an instruction like “give me a mug” may refer to several similar items. In such cases, we annotate the data with all candidate object indices, e.g., \texttt{[2, 7, 18]}, indicating the set of plausible referents.

\end{itemize}

\begin{figure}[b!]
    \centering
    \includegraphics[trim=30 0 30 0, clip, width=0.8\columnwidth]{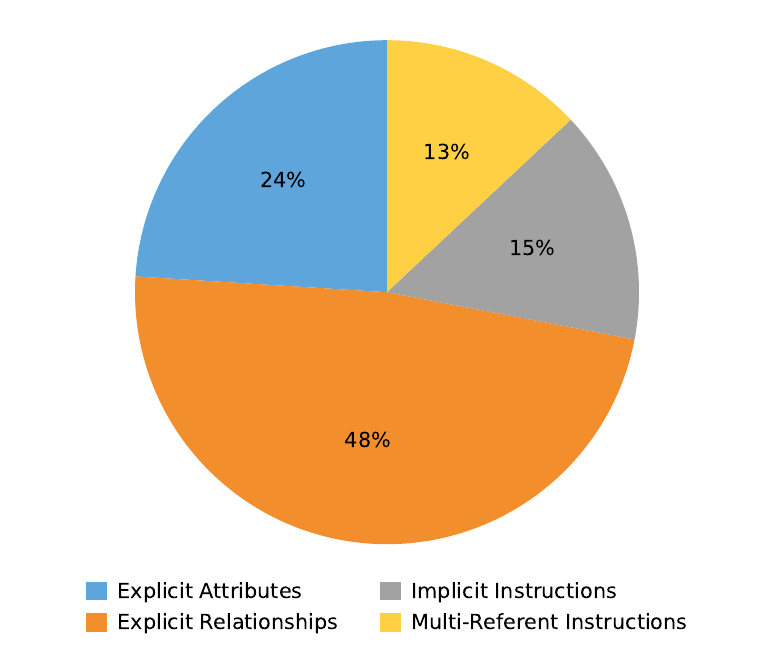}
    \caption{Dataset breakdown by Instruction Types.}
    \label{fig10}
\end{figure}

% PART 2 -------------------
% Instruction Grounding Table

% table multimodal
\begin{table*}[th!]
\centering
\small
\begin{tabular}{lccc ccc ccc}
\toprule
\multirow{2}{*}{Model} & \multicolumn{3}{c}{Easy} & \multicolumn{3}{c}{Medium} & \multicolumn{3}{c}{Hard} \\
 & im & attr & rel & im & attr & rel & im & attr & rel \\
\midrule
VLM-LLM                   & 0.050 & 0.516 & 0.131 & 0.010 & 0.336 & 0.186 & 0.000 & 0.318 & 0.174 \\
Gemini-2.5-Pro            & 0.778 & 0.889 & 0.830 & 0.847 & 0.814 & 0.815 & 0.985 & 0.784 & 0.858 \\
Gemini-2.0                & 0.833 & 0.889 & 0.774 & 0.819 & 0.721 & 0.642 & 1.000 & 0.668 & 0.469 \\
GPT-5-auto                & 0.950 & 1.000 & 0.867 & 0.903 & 0.960 & 0.899 & 0.917 & 0.916 & 0.816 \\
GPT-5-mini                & 0.850 & 1.000 & 0.894 & 0.819 & 0.956 & 0.823 & 0.958 & 0.759 & 0.813 \\
GPT-4.5                   & 0.944 & 0.889 & 0.894 & 0.917 & 0.838 & 0.722 & 0.958 & 0.719 & 0.698 \\
GPT-4o                    & 0.850 & 1.000 & 0.778 & 0.819 & 0.948 & 0.680 & 0.901 & 0.697 & 0.469 \\
o4                        & 0.950 & 1.000 & 0.907 & 0.847 & 0.988 & 0.837 & 0.917 & 0.809 & 0.804 \\
4o-mini                   & 0.750 & 0.717 & 0.550 & 0.764 & 0.771 & 0.596 & 0.750 & 0.382 & 0.248 \\
GPT-4V                    & 0.650 & 0.750 & 0.598 & 0.750 & 0.737 & 0.662 & 0.625 & 0.417 & 0.455 \\
Qwen2-VL                  & 0.800 & 0.917 & 0.830 & 0.792 & 0.756 & 0.738 & 0.875 & 0.700 & 0.529 \\
LLaMA 3.2 Vision 90B      & 0.750 & 0.850 & 0.704 & 0.708 & 0.853 & 0.711 & 0.875 & 0.491 & 0.521 \\
LLaMA 3.2 Vision 90B-Q4   & 0.800 & 0.667 & 0.598 & 0.625 & 0.719 & 0.554 & 0.542 & 0.464 & 0.300 \\
LLaMA 3.2 Vision 11B      & 0.650 & 0.667 & 0.631 & 0.764 & 0.710 & 0.556 & 0.833 & 0.536 & 0.342 \\
LLaMA 3.2 Vision 11B-Q4   & 0.650 & 0.567 & 0.502 & 0.694 & 0.757 & 0.555 & 0.542 & 0.498 & 0.450 \\
\bottomrule
\end{tabular}
\caption{Performance on the complex instruction grounding task. Abbreviations: \textit{im} denotes implicit instruction, \textit{attr} denotes attribute-based instruction, and \textit{rel} denotes relation-based instruction.}

\label{tab:5}
\end{table*}

\vspace{15pt}
A human-in-the-loop process was employed to ensure high-quality data collection.
\begin{itemize}
    \item \textit{Initial Object Identification}: We used GPT-4o to identify objects in an image and referring them by type, explicit attributes, and detailed location relations.
    \item \textit{Human Verification.} The authors of this paper reviewed and modified the outputs to ensure their correctness. 
    \item  \textit{Instruction Generation.} After verification, GPT-4 was tasked with generating simple, clear instructions for different objects.
    \item  \textit{Final Review.} These instructions underwent another round of verification to ensure clarity and accuracy. 
\end{itemize}

This high-quality dataset consisting of 473 instructions, with a detailed breakdown of each instruction type presented in Fig.~\ref{fig10}.
\vspace{15pt}

% fig failure cases 
\begin{figure*}
    \centering
    \includegraphics[width=\linewidth, trim=0 2in 1.2in 0, clip]{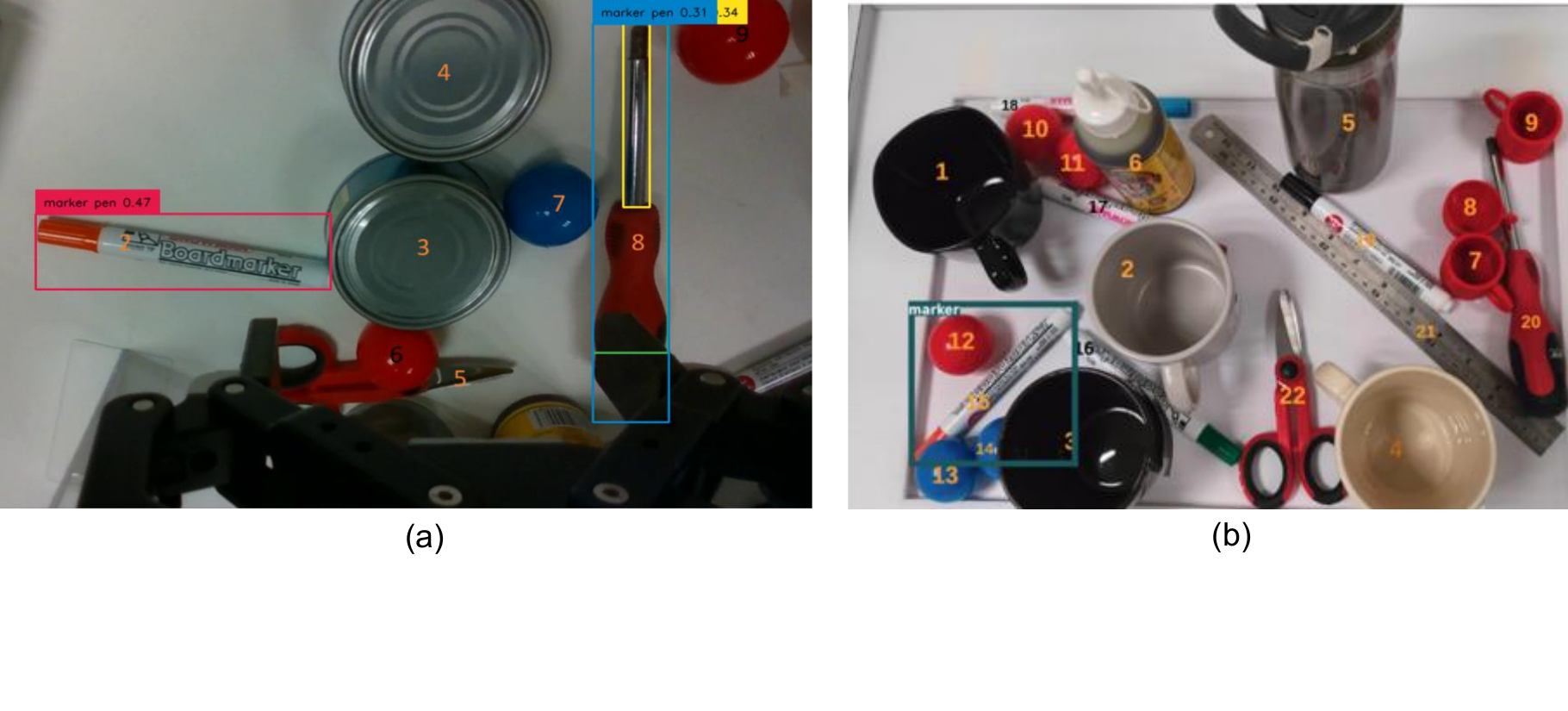}
    \vspace{-20pt}
    \caption{Examples of Instruction Grounding. (a) “the marker on the left”, (b) “the marker aligned with the ruler”.}
    \label{fig11}
\end{figure*}

\begin{figure*}
    \centering
    \includegraphics[width=\linewidth, trim=0 1.2in 3.6in 0, clip]{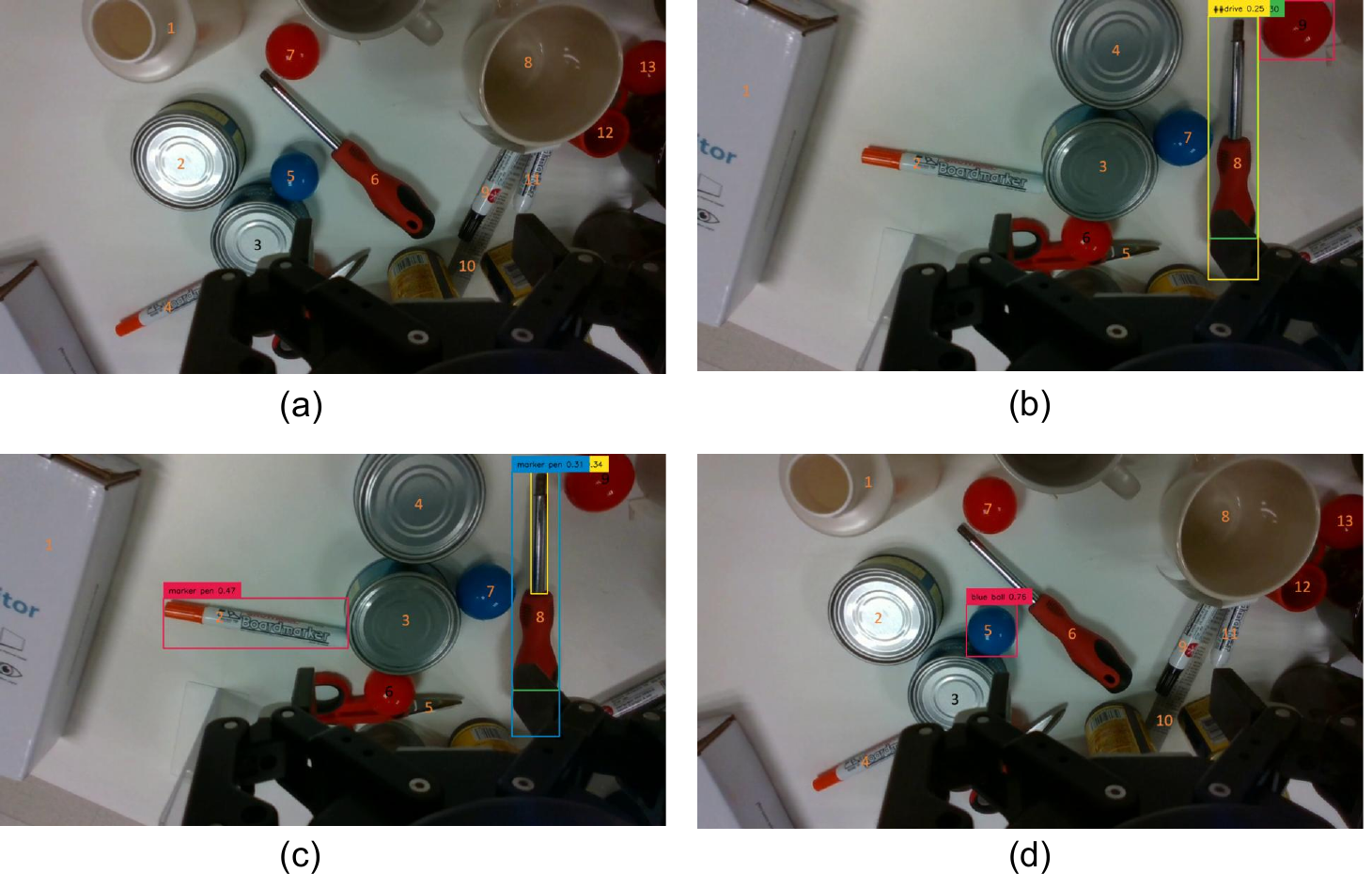}
    \caption{Examples of Object Grounding. (a) ``ball'', (b) ``screwdriver'', (c) ``marker pens'', (d) ``blue ball''.}
    \label{fig12}
\end{figure*}

\section{Grounding Experiments}
\subsection{Complex Instruction Grounding for Goal Specification}
\textit {Cross-modal Disambiguation} represents a particularly challenging component of goal specification. To quantify the model capability in this dimension, we employed attribute-based and relative relationship instructions to uniquely identify a target among multiple candidates.  The goal specification task is formulated as follows:

Given a visual input $I \in \mathbb{R}^ { H \times W \times 3 }$ and an instruction $t$, the objective is to predict the target object according to $o^* = \arg\max_{o \in \mathcal{O}} P(o \,|\, I, t; \theta_{FM})$, where $o^\star \in \mathcal O$ and $\mathcal O$ denotes the set of candidate objects. Models are evaluated using macro-average accuracy metric.

\subsection{Failure Cases of Specialist VLM Pipeline}
Grounding DINO, despite popular for zero-shot detection, is not robust in open scenes. It successfully detected ``blue ball'' while failed to detect ``ball'', indicating its reliance on visual features. Similarly, featureless metal cans pose a great challenge for Grounding DINO, which were almost omitted in the detection results. 

For complex instruction grounding, Grounding DINO and GPT-4 were chained together to ``guess'' the target by the LLM based on the candidate bounding boxes. The failure cases were illustrated in the Fig.~\ref{fig11} and Fig.~\ref{fig12}.

\subsection{Multimodal LLMs Performance}
The performance of Multimodal LLMs on complex grounding across \textsc{Easy}, \textsc{Medium} and \textsc{Hard} groups are shown in Table~\ref{tab:5}.

% PART 3 -------------------
\section{Manipulation Experiments}
OpenVLA and $\pi_0$ were partially fine-tuned on this dataset, while Diffusion Policy (DP) and Action Chunking Transformer (ACT) were trained from scratch. Due to the limited size of our custom dataset, full fine-tuning of RT-1, OpenVLA, SpatialVLA, and NORA was performed using the Open-X-embodiment and LIBERO datasets. We evaluated model performance on both a real-world WidowX robotic platform and the LIBERO simulation benchmark.

\subsection{Fine-tuning details for VLA}
Partial fine-tuning was conducted on a single NVIDIA A6000 GPU (48 GB VRAM) over a period of three days. To ensure a fair comparison, a batch size of 1 was used across all models. The results are presented in Fig.~\ref{fig5}.

\begin{figure*}
    \centering
    %\hspace*{-1.2cm}
    \includegraphics[width=1\linewidth]{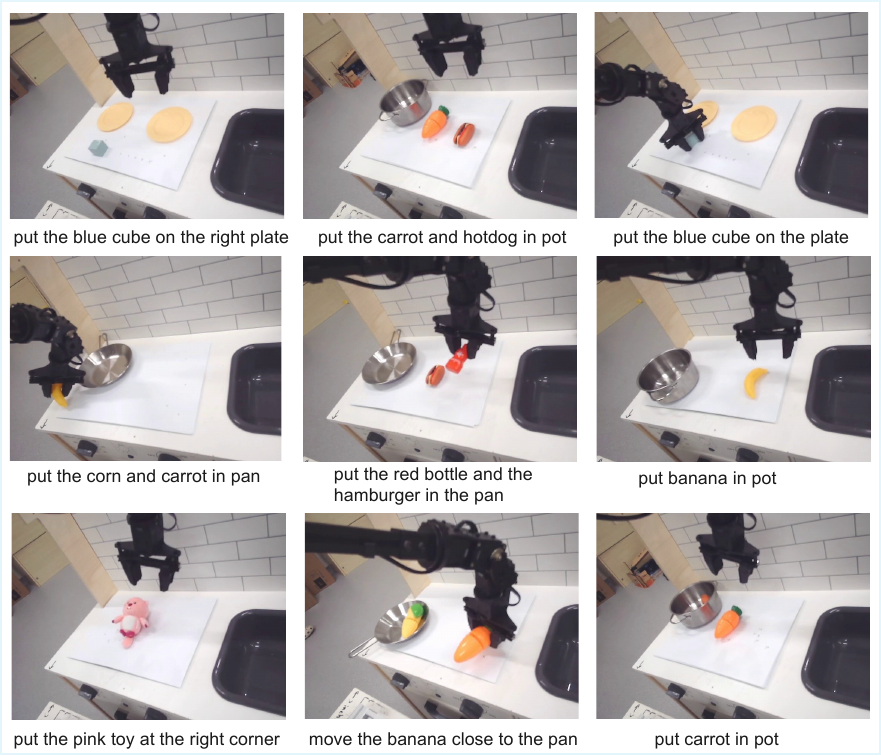}
    \caption{Real-world robot environments and task setups. We evaluate these models across 9 diverse tasks to assess its instruction understanding, spatial reasoning, and multi-task motion planning capabilities.}
    \label{fig:setting}
\end{figure*}

Full fine-tuning of RT-1, OpenVLA, SpatialVLA, and NORA was conducted on a compute node equipped with 8$\times$H100 GPUs. The fine-tuned models were evaluated on 9 diverse real-world manipulation tasks, as shown in Fig.~\ref{fig:setting}. Success rates are summarized in Table~\ref{tab:task_performance}, demonstrating NORA’s superior policy generation capabilities across three task categories: out-of-distribution object grasping, spatial reasoning, and multi-object manipulation. 

\begin{table*}[h]
\centering
\small % or \scriptsize
\renewcommand{\arraystretch}{1.2}
\setlength{\tabcolsep}{5pt} % reduce column padding
\resizebox{\textwidth}{!}{\begin{tabular}{llcccc}
\toprule
\textbf{Category} & \textbf{Task} & \textbf{RT-1} & \textbf{OpenVLA} & \textbf{SpatialVLA} & \textbf{NORA} \\
\midrule
\multirow{3}{*}{Multiple objects} 
& Put the red bottle and the hamburger in the pan & 0 & 20 & 0 & \textbf{40} \\
& Put the carrot and hotdog in pot & 0 & 0 & 0 & \textbf{30} \\
& Put the corn and carrot in the pan & 0 & \textbf{30} & 0 & \textbf{30} \\
\midrule
\multirow{3}{*}{OOD object} 
& put carrot in pot & 0 & 80 & 20 & \textbf{90} \\
& Put banana in pot & 1 & 40 & 0 & \textbf{90} \\
& Put the blue cube on the plate & 0 & 50 & 0 & \textbf{70} \\
\midrule
\multirow{3}{*}{Spatial} 
& Put the pink toy at the right corner & 0 & \textbf{60} & 30 & \textbf{60} \\
& Put the blue cube on the right plate & 0 & \textbf{30} & 0 & 20 \\
& Move the banana close to the pan & 30 & 50 & 50 & \textbf{80} \\
\midrule
\multicolumn{2}{l}{\textbf{Average}} & 4.4 & 40 & 11.1 & \textbf{56.7} \\
\bottomrule
\end{tabular}}
\caption{Task performance comparison across different categories and models.}
\label{tab:task_performance}
\end{table*}

\subsection{Impact of Action Chunking}
\subsubsection{Action Chunking Performs on WidowX.} 
To investigate the effectiveness of action chunking, we selected NORA-LONG and SpatialVLA for evaluation. Tasks were chosen from three categories: (1) “\texttt{put the carrot in the pot},” (2) “\texttt{put the red bottle and hamburger in the pot},” and (3) “\texttt{put the pink toy at the right corner}.” In initial experiments, all predicted actions (5 actions for NORA-LONG, 4 actions for SpatialVLA) were executed sequentially without replanning. This frequently caused the WidowX robot to crash into the environment due to the accumulation of overly large movements.

Subsequently, we modified the execution policy to only perform the first action in each predicted chunk. This adjustment resolved the collision issue but the model performance is still degraded.

\subsubsection{Action chunking improves performance in simulation.} We hypothesize that action chunking is more effective at higher control frequencies. For example, Diffusion Policy generates commands at 10 Hz, which are then interpolated to 125 Hz for execution. Similarly, OpenVLA-OFT+ employs action chunking and shows improved performance in real-world ALOHA tasks, which run at 25 Hz.

Since our real robotic platforms do not support high-frequency control, we tested this hypothesis in the LIBERO simulation environment (20 Hz). We fine-tuned both NORA and NORA-LONG on this benchmark with an action chunk size of 5, producing two variants: NORA-finetuned-AC and NORA-Long-finetuned.

Results show that NORA-finetuned-AC significantly outperforms NORA-finetuned across all LIBERO benchmarks, with a higher average success rate. Notably, NORA-Long-finetuned outperforms all baseline models (see Table~\ref{tab:4}), highlighting the benefits of pretraining with action chunking and its transferability to long-horizon tasks. However, it is important to note that LIBERO is a simulation environment and may not reflect real-world performance at high control frequencies.
\begin{figure}[t]
    \centering
    \includegraphics[width=\linewidth]{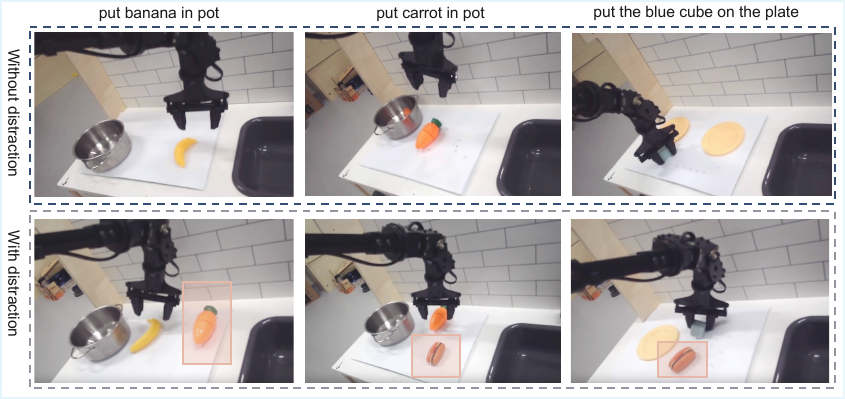}
    \caption{Comparison of tasks with and without distraction.}
    \label{fig:disturbance}
\end{figure}

\begin{table}[h]
    \centering
    \resizebox{\linewidth}{!}{\begin{tabular}{lcc}
        \toprule
        \textbf{TASK} & \textbf{OpenVLA} & \textbf{NORA} \\
        \midrule
        \textbf{without Distraction} & & \\
        put carrot in pot & 8 & 9 \\
        put banana in pot & 4 & 9 \\
        put the blue cube on the plate & 5 & 7 \\
        \midrule
        \textbf{with Distraction} & & \\
        put carrot in pot & 6 & 8 \\
        put banana in pot & 6 & 4 \\
        put the blue cube on the plate & 3 & 5 \\
        \bottomrule
    \end{tabular}}
    \caption{Comparison of task performance between OpenVLA and NORA under conditions with and without distraction. Each value denotes the number of successful executions out of 10 trials.}
    \label{tab:disturbance}
\end{table}

\subsection{Robustness to Disturbance}
To evaluate robustness, we selected three straightforward tasks (shown in Fig.~\ref{fig:disturbance}) and introduced distractor objects into the environment. Initially, both OpenVLA and NORA performed well. However, their success rates declined significantly with the introduction of distractions. This highlights the sensitivity of current VLA models to out-of-distribution disturbances. The average success rates across the three tasks are presented in Table~\ref{tab:distractors}, while the detailed number of successful executions out of 10 trials is summarized in Table~\ref{tab:disturbance}.

\begin{table}
\centering
\caption{Average Success Rate (\%) without (w/o) and with (w/) Distractors}
\label{tab:distractors}
\begin{tabular}{lcc}
\toprule
\text{Model} & \text{w/o Distractors} & \text{w/ Distractors} \\
\midrule
OpenVLA & 56.7 & 50 \\
NORA    & 83.3 & 56.7 \\
\bottomrule
\end{tabular}
\end{table}

% PART 4 -------------------

\section{Modular Claw Machine Prototype}
To facilitate the evaluation of different VLMs in robotic manipulation, we developed a voice-controlled testbed using a UR5 robotic arm\footnote{\url{https://github.com/HRItdy/claw_machine}}. 
The system architecture, shown in Fig.~\ref{fig:diagram}, comprises the following five modules:
\begin{itemize}[leftmargin=1em, itemsep=0pt, topsep=0pt]
\item \textbf{Speech Transcription:} Powered by Microsoft Azure's speech recognition service.

\item \textbf{Task Decomposition:} Based on GPT-3.5 and GPT-4 using prompting paradigms adapted from ChatGPT for Robotics.
\item \textbf{Object Detection:} Utilizes GroundingDINO and OWL-ViT for object detection.

\item \textbf{Object Segmentation:} Employs Segment Anything Model (SAM) and FastSAM for segmenting detected objects.

\begin{figure}[hb]
    \centering
    \includegraphics[width=\linewidth]{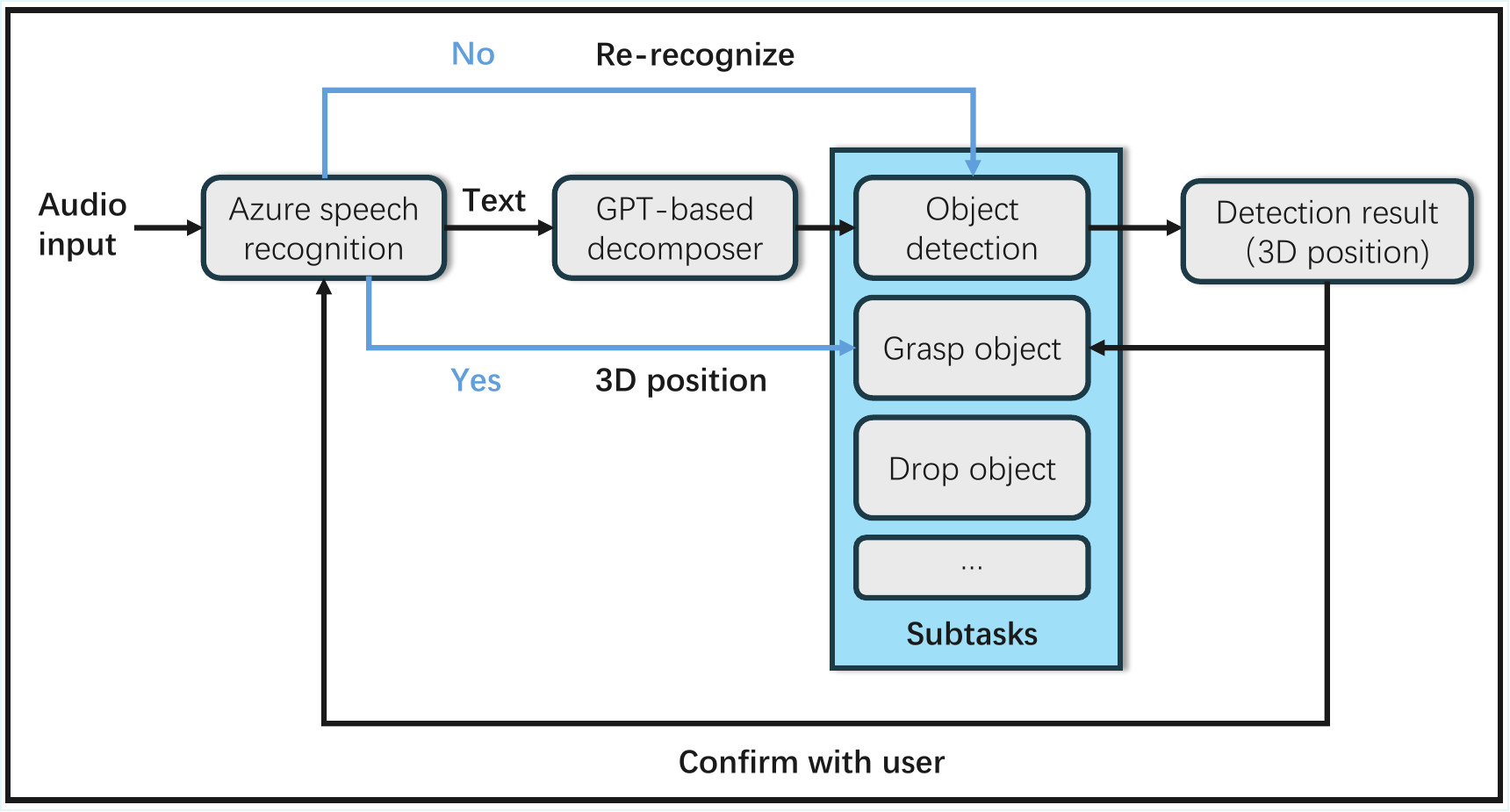}
    \caption{The system architecture of the testbed for VLMs.}
    \label{fig:diagram}
\end{figure}
\item \textbf{Manipulation:} Low-level actions are generated by GraspAnything or GraspNet.
\end{itemize}

This modular testbed enables rapid integration and benchmarking of different models within a real robotic system.

\end{document}